\documentclass[10pt,twocolumn,letterpaper]{article}

\usepackage{cvpr}
\usepackage{times}
\usepackage{epsfig}
\usepackage{graphicx}
\usepackage{amsmath}
\usepackage{amssymb}
\usepackage{multirow}
\usepackage{slashbox}

\usepackage{subcaption}
\DeclareCaptionFont{tiny}{\tiny}
\usepackage[pagebackref=true,breaklinks=true,letterpaper=true,colorlinks,bookmarks=false]{hyperref}

\usepackage[breaklinks=true,bookmarks=false]{hyperref}

\cvprfinalcopy 

\def\cvprPaperID{****} 

\ifcvprfinal\fi
\begin{document}

\title{Stochastic Attraction-Repulsion Embedding for Large Scale Image Localization}

\author{Liu Liu $^{1,2}$, Hongdong Li $^{1,2}$ and Yuchao Dai $^3$\\
$^{1}$ Australian National University, Canberra, Australia  \\
$^{2}$ Australian Centre for Robotic Vision \\
$^{3}$ School of Electronics and Information, Northwestern Polytechnical University, Xian, China \\
\tt\small{ \{Liu.Liu; hongdong.li\}}@anu.edu.au; daiyuchao@nwpu.edu.cn
}

\maketitle

\begin{abstract}

This paper tackles the problem of large-scale image-based localization (IBL) where the spatial location of a query image is determined by finding out the most similar {\em reference images} in a large database.  For solving this problem, a critical task is to learn discriminative image representation that captures informative information relevant for localization. We propose a novel representation learning method having higher location-discriminating power. 

It provides the following contributions: 1) we represent a place (location) as a set of exemplar images depicting the
same landmarks and aim to maximize similarities among intra-place images while minimizing similarities among inter-place images; 2) we model a similarity measure as a probability distribution on $L_2$-metric distances between intra-place and inter-place image representations; 3) we propose a new Stochastic Attraction and Repulsion Embedding (SARE) loss function minimizing the KL divergence between the learned and the actual probability distributions; 4) we give theoretical comparisons between SARE, triplet ranking \cite{arandjelovic2016netvlad} and contrastive losses \cite{radenovic2016cnn}. It provides insights into why SARE is better by analyzing gradients.

Our SARE loss is easy to implement and pluggable to any CNN. Experiments show that our proposed method improves the localization performance on standard benchmarks by a large margin.  Demonstrating the broad applicability of our method, we obtained the $3^{rd}$ place out of 209 teams in the 2018 Google Landmark Retrieval Challenge \cite{Google_Landmark}. Our code and model are available at \url{https://github.com/Liumouliu/deepIBL}.

\end{abstract}

\section{Introduction}

\begin{figure}
\centering
\includegraphics[width=0.485\textwidth]{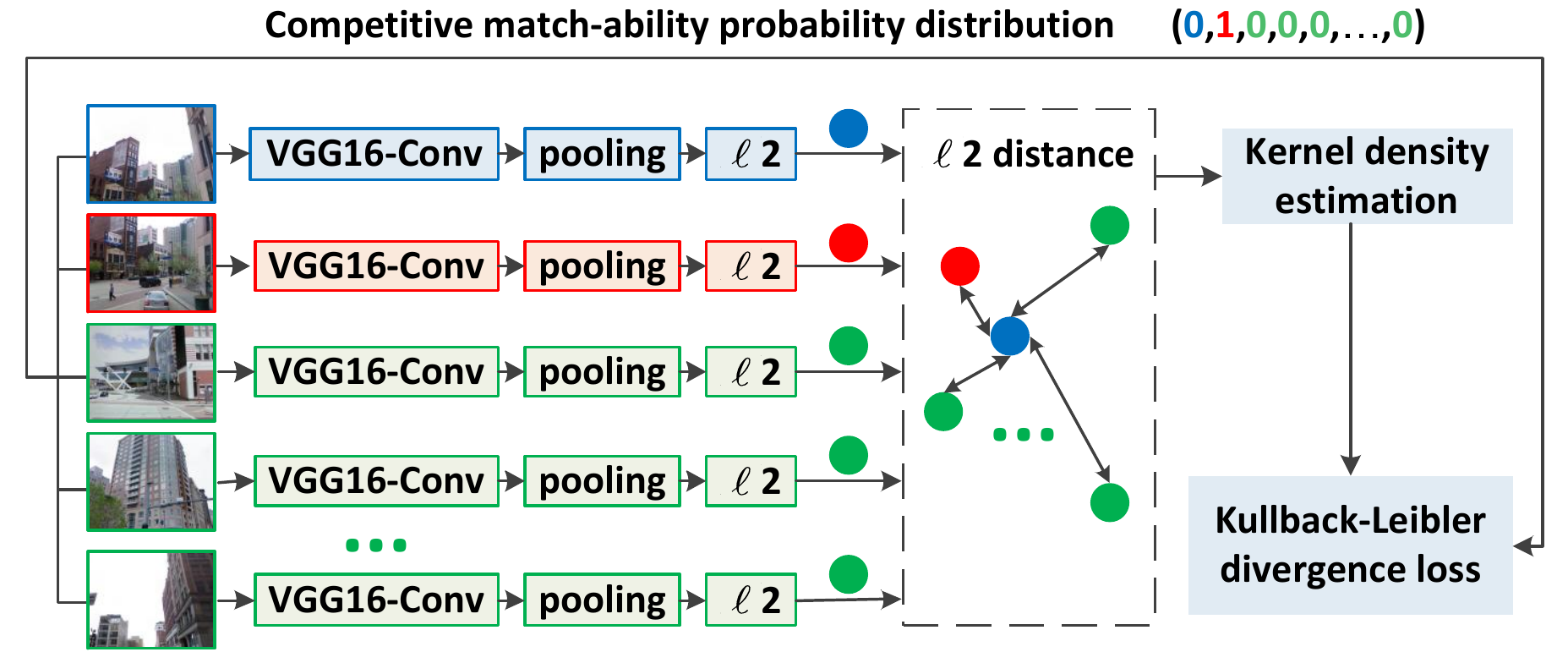}
\caption{\small{The pipeline of our method. We use the VGG16 net \cite{simonyan2014very} with only convolution layers as our architecture. NetVLAD \cite{arandjelovic2016netvlad} pooling is used to obtain compact image representations. The feature vectors are post $L_2$ normalized. The $L_2$ distance between the \textcolor{blue}{query}-\textcolor{red}{positive} and the \textcolor{blue}{query}-\textcolor{green}{negative} images are calculated, and converted to a probability distribution. The estimated probability distribution is compared with the ground-truth match-ability distribution, yielding the Kullback-Leibler divergence loss.
}}
\label{fig:pipeline}
\end{figure}

The task of Image-Based Localization (IBL) is to estimate the geographic location of where a query image is taken, based on comparing it against geo-tagged images from a city-scale image database (\ie a map). IBL has attracted considerable attention recently due to the wide-spread potential applications such as in robot navigation \cite{mur2015orb} and  VR/AR \cite{middelberg2014scalable,ventura2014global}.  Depending on whether or not 3D point-clouds are used in the map, existing IBL methods can be roughly classified into two groups: {\em image-retrieval} based methods \cite{arandjelovic2016netvlad,kim2017crn,sattler2017large,Noh_2017_ICCV,Vo_2017_ICCV,radenovic2016cnn} and {\em direct 2D-3D matching} based methods \cite{sattler2011fast,sattler2017efficient,li2010location,Liu_2017_ICCV,CarlSematic}. 

This paper belongs to the {\em image-retrieval} group for its effectiveness at large
scale and robustness to changing conditions \cite{sattler2017benchmarking}. For {\em image-retrieval} based methods, the main challenge is how to discriminatively represent images so that images depicting same landmarks would have similar representations while those depicting different landmarks would have dissimilar representations. The challenge is underpinned by the typically large-scale image database, in which many images may contain repetitive structures and similar landmarks, causing severe ambiguities. 

Convolution Neural Networks (CNNs) have demonstrated great success for the IBL task \cite{arandjelovic2016netvlad,kim2017crn,Noh_2017_ICCV,gordo2016deep,gordo2017end,radenovic2016cnn}. Typically, CNNs trained for image classification task are fine-tuned for IBL. As far as we know, all the state-of-the-art IBL methods focus on how to effectively aggregate a CNN feature map to obtain discriminative image representation, but have overlooked another important aspect which can potentially boost the IBL performance markedly. The important aspect is how to effectively organize the aggregated image representations. So far, all state-of-the-art IBL methods use triplet ranking and contrastive embedding to supervise the representation organization process.

This paper fills this gap by proposing a new method to effectively organize the image representations (embeddings). We first define a ``place'' as a set of images depicting same location landmarks, and then directly enforce the intra-place image similarity and inter-place dissimilarity in the embedding space. Our goal is to cluster learned embeddings from the same place while separating embeddings from different places. Intuitively, we are organizing image representations using places as agents. 

The above idea may directly lead to a multi-class classification problem if we can label the ``place'' tag for each image. Apart from the time-consuming labeling process, the formulation will also result in too many pre-defined classes and we need a large training image set to train the classification CNN net. Recently-proposed methods \cite{Vo_2017_ICCV,weyand2016planet} try to solve the multi-class classification problem using large GPS-tagged training dataset. In their setting, a class is defined as images captured from nearby geographic positions while disregarding their visual appearance information. Since images within the same class do not necessarily depict same landmarks, CNN may only learn high-level information \cite{Vo_2017_ICCV} for each geographic position, thus inadequate for accurate localization.

Can we capture the intra-place image ``attraction'' and inter-place image ``repulsion'' relationship with limited data? To tackle the ``attraction'' and ``repulsion'' relationship, we formulate the IBL task as image similarity-based binary classification in feature embedding space. Specifically, the similarity for images in the same place is defined as 1, and 0 otherwise. This binary-partition of similarity is used to capture the intra-place ``attraction'' and inter-place ``repulsion''. To tackle the limited data issue, we use triplet images to train CNN, consisting of one query, positive (from the same place as the query), and negative image (from a different place). Note that a triplet is a minimum set to define the intra-place ``attraction'' and inter-place ``repulsion''. 

Our CNN architecture is given in Fig.~\ref{fig:pipeline}. We name our metric-learning objective as Stochastic Attraction and Repulsion Embedding (SARE) since it captures pairwise image relationships under the probabilistic framework. Moreover, our SARE objective can be easily extended to handle multiple negative images coming from different places, \ie enabling competition with multiple other places for each place. 
In experiments, we demonstrate that, with SARE, we obtain improved performance on various IBL benchmarks. Validations on standard image retrieval benchmarks further justify the superior generalization ability of our method.

\section{Related Work}
There is a rich family of work in IBL. We briefly review CNN-based image representation learning methods. Please refer to \cite{DBLP:journals/corr/Wu16e, zheng2018sift} for an overview.

While there have been many works \cite{RazavianBaseline,gordo2016deep,gordo2017end,radenovic2016cnn,arandjelovic2016netvlad,Noh_2017_ICCV,kim2017crn,sattler2017large} in designing effective CNN feature map aggregation methods for IBL, they almost all exclusively use triplet or contrastive embedding objective to supervise CNN training. Both of these two objectives in spirit pulling the $L_2$ distance of matchable image pair while pushing the $L_2$ distance of non-matching image pair. While they are effective, we will show that our SARE objective outperforms them in the IBL task later. Three interesting exceptions which do not use triplet or contrastive embedding objective are the planet \cite{weyand2016planet}, IM2GPS-CNN \cite{Vo_2017_ICCV}, and CPlaNet \cite{seo2018cplanet}. They formulate IBL as a geographic position classification task. They first partition a 2D geographic space into cells using GPS-tags and then define a class per-cell. CNN training process is supervised by the cross-entropy classification loss which penalizes incorrectly classified images. We also show that our SARE objective outperforms the multi-class classification objective in the IBL task.

Although our SARE objective enforces intra-place image ``attraction'' and inter-place image ``repulsion'', it differs from traditional competitive learning methods such as Self-Organizing Map \cite{kohonen1998self} and Vector Quantization \cite{munoz2002expansive}. They are both devoted to learning cluster centers to separate original vectors. No constraints are imposed on original vectors. Under our formulation, we directly impose the ``attraction-repulsion'' relationship on original vectors to supervise the CNN learning process.

\section{Problem Definition and Method Overview}
Given a large geotagged image database, the IBL task is to estimate the geographic position of a query image $q$. Image-retrieval based method first identifies the most visually similar image from the database for $q$, and then use the location of the database image as that of $q$. If the identified most similar image comes from the same place as $q$, then we deem that we have successfully localized $q$, and the most similar image is a positive image, denoted as $p$. If the identified most similar image comes from a different place as $q$, then we have falsely localized $q$, and the most similar image is a negative image, denoted as $n$. 

Mathematically, an image-retrieval based method is executed as follows: First, query image and database images are converted to compact representations (vectors). This step is called image feature embedding and is done by a CNN network. For example, query image $q$ is converted to a fixed-size vector $f_\theta(q)$, where $f$ is a CNN network and $\theta$ is the CNN weight. Second, we define a similarity function $S(\cdot)$  on pairwise vectors. For example, $S\left ( f_\theta(q), f_\theta(p)\right )$ takes vectors $f_\theta(q)$ and $f_\theta(p)$, and outputs a scalar value describing the similarity between $q$ and $p$. Since we are comparing the entire large database to find the most similar image for $q$, $S(\cdot)$ should be simple and efficiently computed to enable fast nearest neighbor search. A typical choice for $S(\cdot)$ is the $L_2$-metric distance, or functions monotonically increase/decrease with the $L_2$-metric distance. 

Relying on feature vectors extracted by un-trained CNN to perform nearest neighbor search would often output a negative image $n$ for $q$. Thus, we need to train CNN using easily obtained geo-tagged training images (Sec.\ref{sec::implentation}). The training process in general defines a loss function on CNN extracted feature vectors, and use it to update the CNN weight $\theta$. State-of-the-art triplet ranking loss (Sec.\ref{sec::Revisiting}) takes triplet training images $q,p,n$, and imposes that $q$ is more similar to $p$ than $n$. Another contrastive loss (Sec.\ref{sec::Contrastive}) tries to separate $q\sim n$ pair by a pre-defined distance margin (see Fig.\ref{fig:triplet_constrastive}). While these two losses are effective, we construct our metric embedding objective in a substantially different way.  

Given triplet training images $q,p,n$, we have the prior knowledge that $q\sim p$ pair is matchable and $q\sim n$ pair is non-matchable. This simple match-ability prior actually defines a probability distribution. For $q\sim p$ pair, the match-ability is defined as 1. For $q\sim n$ pair, the match-ability is defined as 0. Can we respect this match-ability prior in feature embedding space? Our answer is yes. To do it, we directly fit a kernel on the $L_2$-metric distances of $q\sim p$ and $q\sim n$ pairs and obtain a probability distribution. Our metric-learning objective is to minimize the Kullback-Leibler divergence of the above two probability distributions (Sec.\ref{sec::SNE}).

What's the benefit of respecting the match-ability prior in feature embedding space? Conceptually, in this way, we capture the intra-place (defined by $q\sim p$ pair) ``attraction''  and inter-place (defined by $q\sim n$ pair) ``repulsion'' relationship in feature embedding space. Potentially, the ``attraction'' and ``repulsion'' relationship balances the embedded positions of the entire image database well. Mathematically, we use gradients of the resulting metric-learning objective with respect to triplet images to figure out the characteristics, and find that our objective adaptively adjusts the force (gradient) to pull the distance of $q\sim p$ pair, while pushing the distance of $q\sim n$ pair (Sec.\ref{sec::Gradients}). 

\section{
Deep Metric Embedding Objectives in IBL}\label{sec::Revisiting}
In this section, we first give the two widely-used deep metric embedding objectives in IBL - the triplet ranking and contrastive embedding, and they are facilitated by minimizing the triplet ranking and contrastive loss, respectively. We then give our own objective - Stochastic Attraction and Repulsion Embedding (SARE).

\subsection{Triplet Ranking Loss}\label{sec::Revisiting}
The triplet ranking loss is defined by
{
\small
\begin{equation} \label{eq::triplet_violating}
L_\theta\left (q,p,n  \right )=\max\left ( 0,m+ \left \| f_\theta(q)- f_\theta(p)\right \|^2 - \left \| f_\theta(q)- f_\theta(n)\right \|^2\right),
\end{equation}
}
where $m$ is an empirical margin, typically $m=0.1$ \cite{netVLAD_pami,arandjelovic2016netvlad,gordo2016deep,radenovic2016cnn}. $m$ is used to prune out triplet images with $\left \| f_\theta(q)- f_\theta(n)\right \|^2 > m+ \left \| f_\theta(q)- f_\theta(p)\right \|^2$.

\subsection{Contrastive Loss} \label{sec::Contrastive}
The contrastive loss imposes constraint on image pair $i\sim j$ by:
\begin{equation} \label{Eq:ContrastiveLoss}
\begin{split}
L_\theta\left (i,j  \right ) = & \frac{1}{2} \eta\left \| f_\theta(i)- f_\theta(j)\right \|^2 + \\
& \frac{1}{2} (1-\eta)\left(\max\left ( 0,\tau- \left \| f_\theta(i)- f_\theta(j)\right \|\right )^2\right) 
\end{split}
\end{equation}
where for $q\sim p$ pair, $\eta = 1$, and for $q\sim n$ pair, $\eta = 0$. $\tau$ is an empirical margin to prune out negative images with $\left \| f_\theta(i)- f_\theta(j)\right \| > \tau$. Typically, $\tau = 0.7$ \cite{radenovic2016cnn}. 

\begin{figure}
\centering
\includegraphics[width=0.485\textwidth]{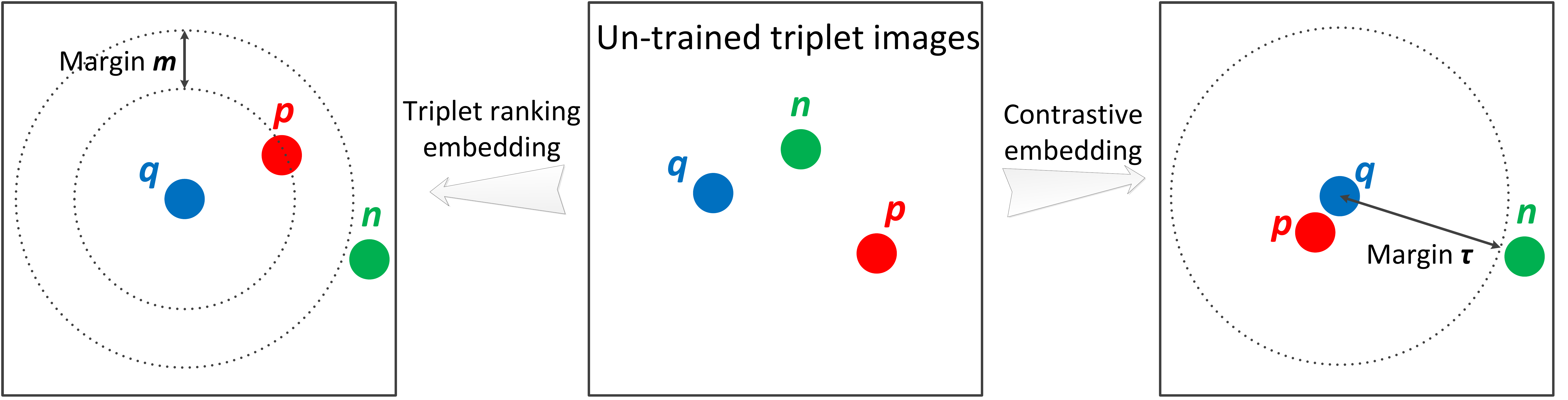}
\caption{\small{ Triplet ranking loss imposes the constraint $\left \| f_\theta(q)- f_\theta(n)\right \|^2 > m+ \left \| f_\theta(q)- f_\theta(p)\right \|^2$. Contrastive loss pulls the $L_2$ distance of $q\sim p$ pair to infinite-minimal, while pushing the $L_2$ distance of $q\sim n$ pair to at least $\tau$-away.
}}
\label{fig:triplet_constrastive}
\end{figure}

Intuitions to the above two losses are compared in Fig.\ref{fig:triplet_constrastive}.

\subsection{SARE-Stochastic Attraction and Repulsion Embedding}\label{sec::SNE}

In this subsection, we present our Stochastic Attraction and Repulsion Embedding (SARE) objective, which is optimized to
learn discriminative embeddings for each ``place''. A triplet images $q,p,n$ define two places, one defined by $q\sim p$ pair and the other defined by $n$. The intra-place and inter-place similarity are defined in a probabilistic framework.  

Given a query image $q$, 
the probability $q$ picks $p$ as its match is conditional probability $h_{p|q}$, which equals to 1 based on the co-visible or matchable prior. The conditional probability $h_{n|q}$ equals to 0 following above definition. Since we are interested in modeling pairwise similarities, we set $h_{q|q} = 0$. Note that the triplet probabilities $h_{q|q}, h_{p|q}, h_{n|q}$ actually define a probability distribution (summing to 1).

In the feature embedding space, we would like CNN extracted feature vectors to respect the above probability distribution. We define another probability distribution $c_{q|q}, c_{p|q}, c_{n|q}$ in the embedding space, and try to minimize the mismatch between the two distributions. The Kullback-Leibler divergence is employed to describe the cross-entropy loss and is given by:
\begin{equation} \label{triplet_loss}
\begin{split}
L_\theta\left (q,p,n  \right ) &=  h_{p|q}\log\left ( \frac{h_{p|q}}{c_{p|q}} \right ) +h_{n|q}\log\left ( \frac{h_{n|q}}{c_{n|q}} \right ) \\
& =-\log\left ( c_{p|q} \right ),
\end{split}
\end{equation}

In order to define the probability $q$ picks $p$ as its match in the feature embedding space, we fit a kernel on pairwise $L_2$-metric feature vector distances. We use three typical-used kernels to compare their effectiveness: Gaussian, Cauchy, and Exponential kernels.
In next paragraphs, we use the Gaussian kernel to demonstrate our method. Loss functions defined by using Cauchy and Exponential kernels are given in Appendix.

For the Gaussian kernel, we have:
{\small
\begin{align} 
c_{p|q} &= \frac{\exp\left ( -\left \| f_\theta(q)- f_\theta(p)\right \|^2 \right )}{\exp\left ( -\left \| f_\theta(q)- f_\theta(p)\right \|^2 \right )+\exp\left ( -\left \| f_\theta(q)- f_\theta(n)\right \|^2 \right )}. \label{Eq:Gaussian:cpq} 
\end{align}
}
In the feature embedding space, the probability of $q$ picks $n$ as its match is given by $c_{n|q} = 1 - c_{p|q}$. If the embedded feature vectors $f_\theta(q)$ and $f_\theta(p)$ are sufficiently near, and $f_\theta(q)$ and $f_\theta(n)$ are far enough under the $L_2$-metric, the conditional probability distributions $c_{\cdot|q}$ and $h_{\cdot|q}$ will be equal. Thus, our SARE objective aims to find an embedding function $f_\theta(\cdot)$ that pulls the $L_2$ distance of $f_\theta(q) \sim f_\theta(p)$ to infinite-minimal, and that of $f_\theta(q) \sim f_\theta(n)$ to infinite-maximal.

Note that although ratio-loss \cite{hoffer2015deep} looks similar to our Exponential kernel $\exp(-||x-y||)$ defined loss function, they are theoretically different.  The building block of ratio-loss is $\exp(||x-y||)$, and it directly applies $\exp()$ to distance $||x-y||$. This is problematic since it is not positive-defined (Please refer to Proposition 3\&4 \cite{scholkopf2001kernel} or \cite{schoenberg1938metric}). 

\section{Comparing the Three Losses} \label{sec::Gradients}
In this section, we illustrate the connections between the above three different loss functions. This is approached by deriving and comparing their gradients,  which are key to the back-propagation stage in networks training.  Note that gradient may be interpreted as the resultant force created by a set of springs between image pair \cite{maaten2008visualizing}.   For the gradient with respect to the positive image $p$, the spring pulls the $q\sim p$ pair. For the gradient with respect to the negative image $n$, the spring pushes the $q\sim n$ pair. 

\begin{figure*}
\begin{center}
\includegraphics[width=0.185\textwidth]{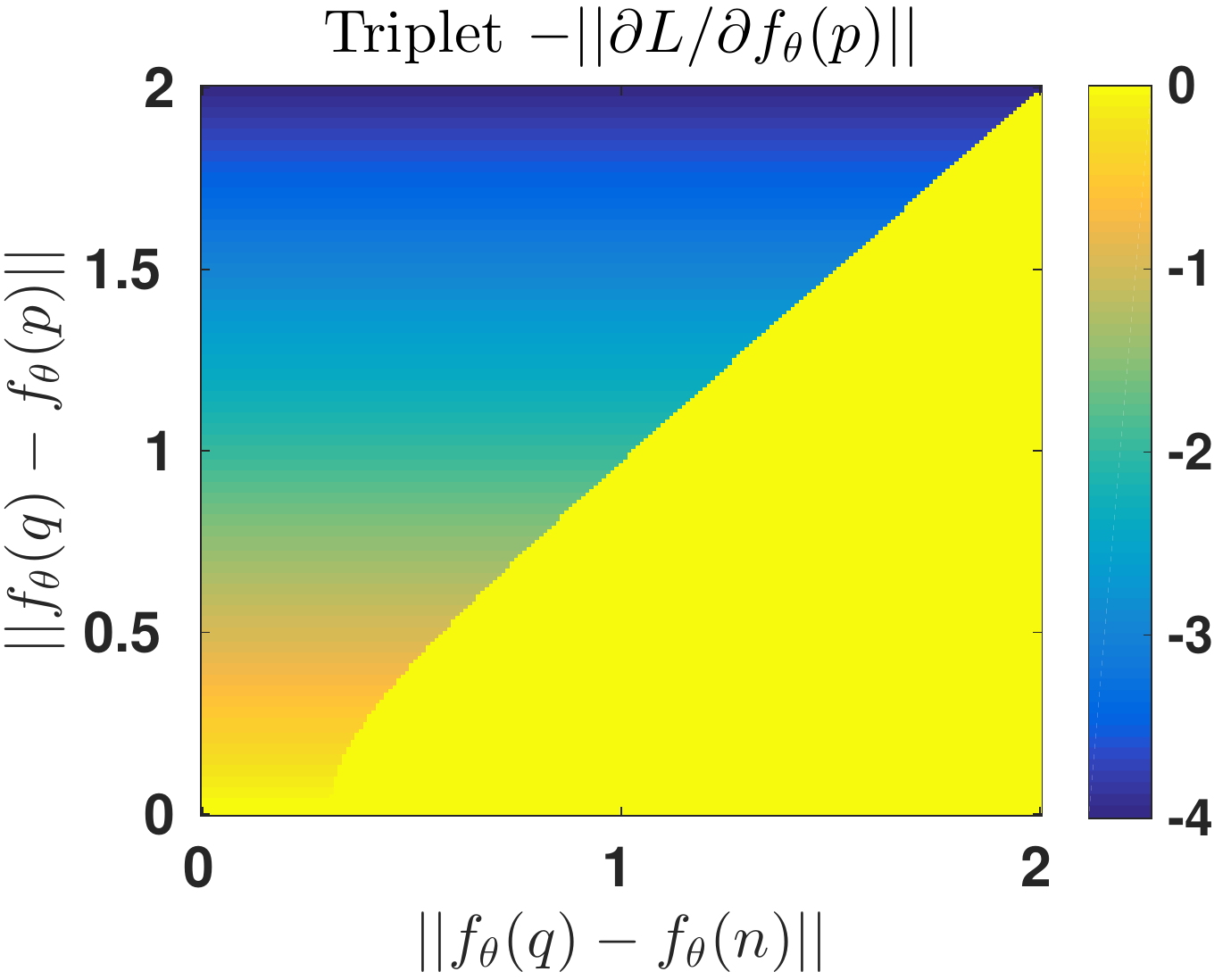}
\includegraphics[width=0.195\textwidth]{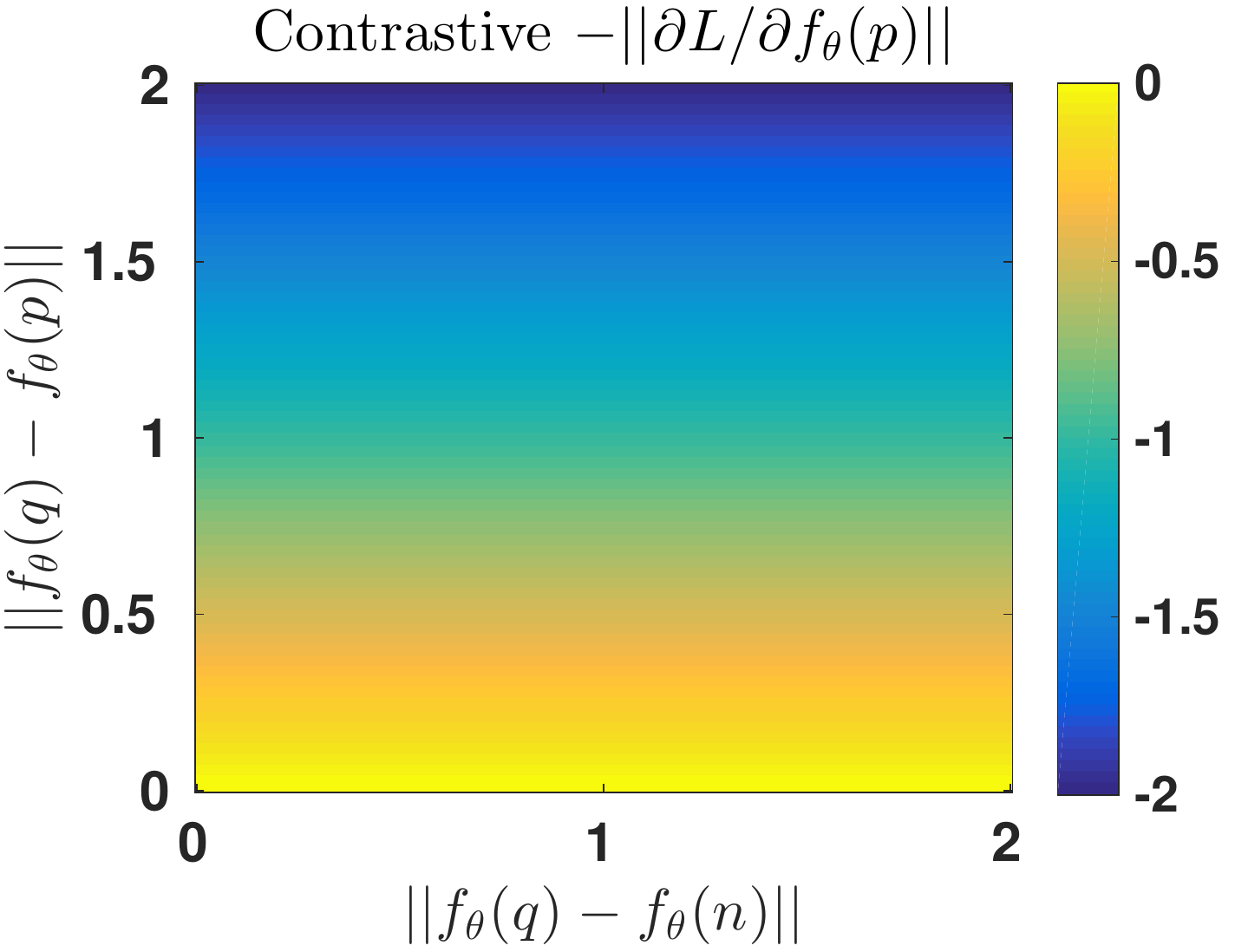}
\includegraphics[width=0.195\textwidth]{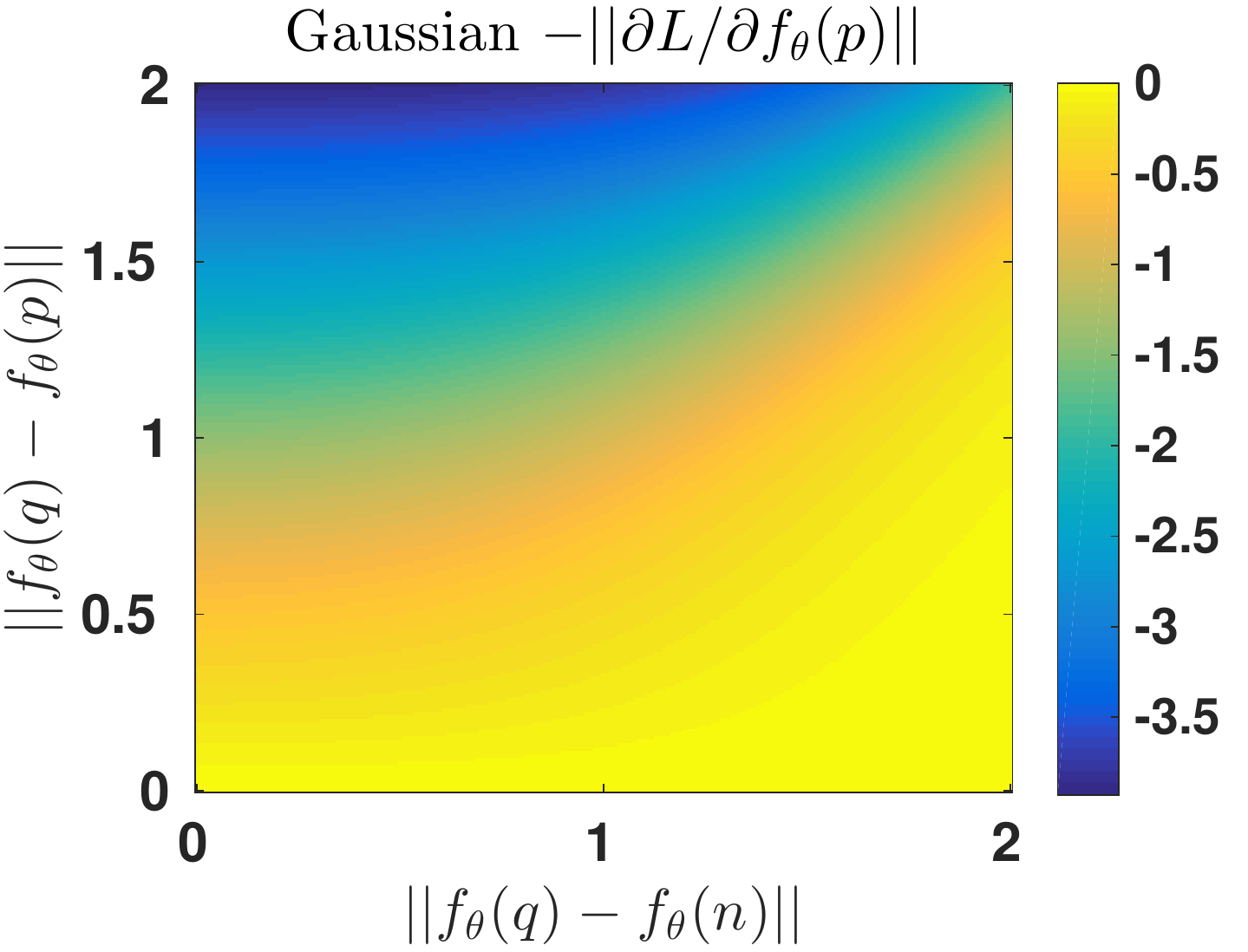}
\includegraphics[width=0.195\textwidth]{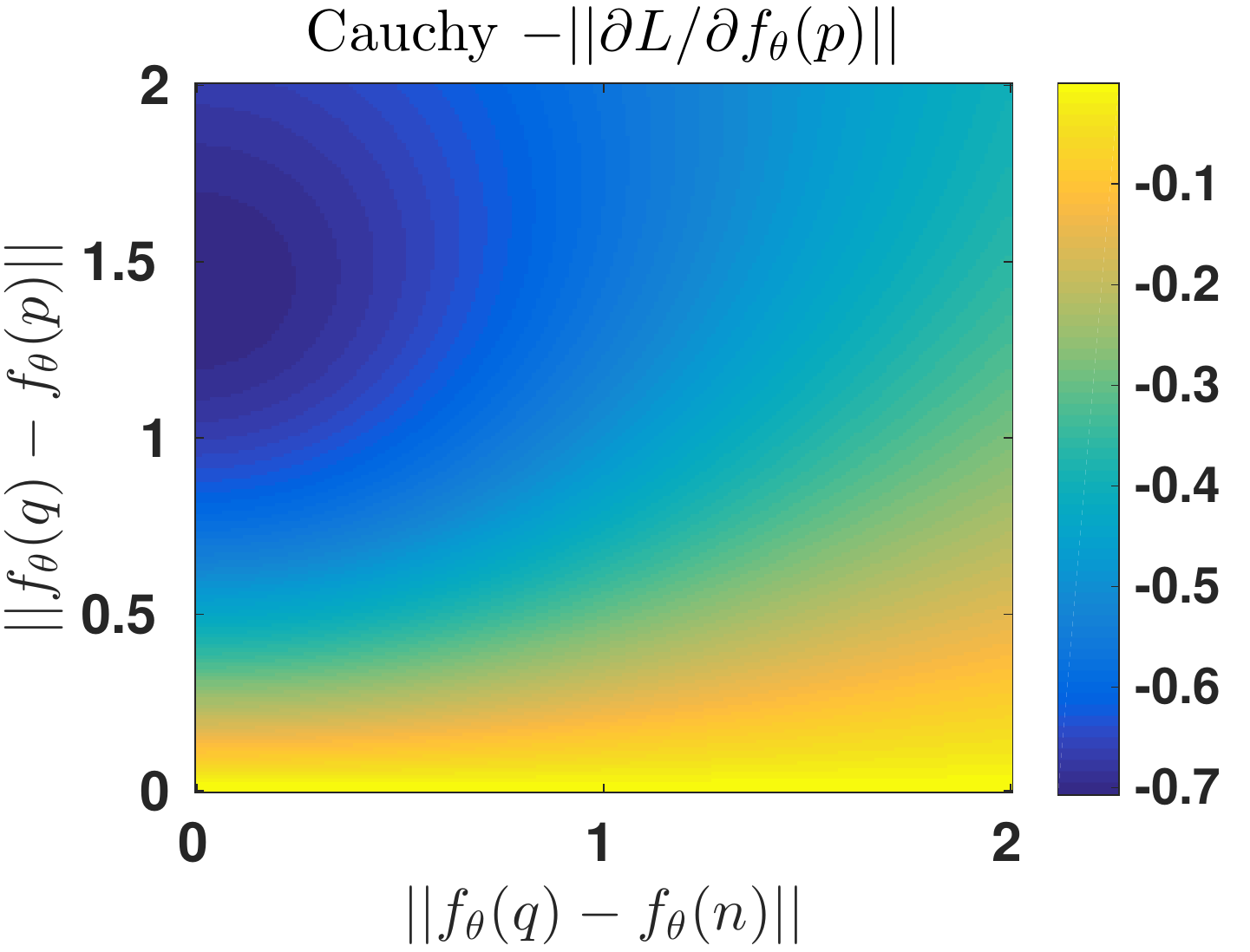}
\includegraphics[width=0.195\textwidth]{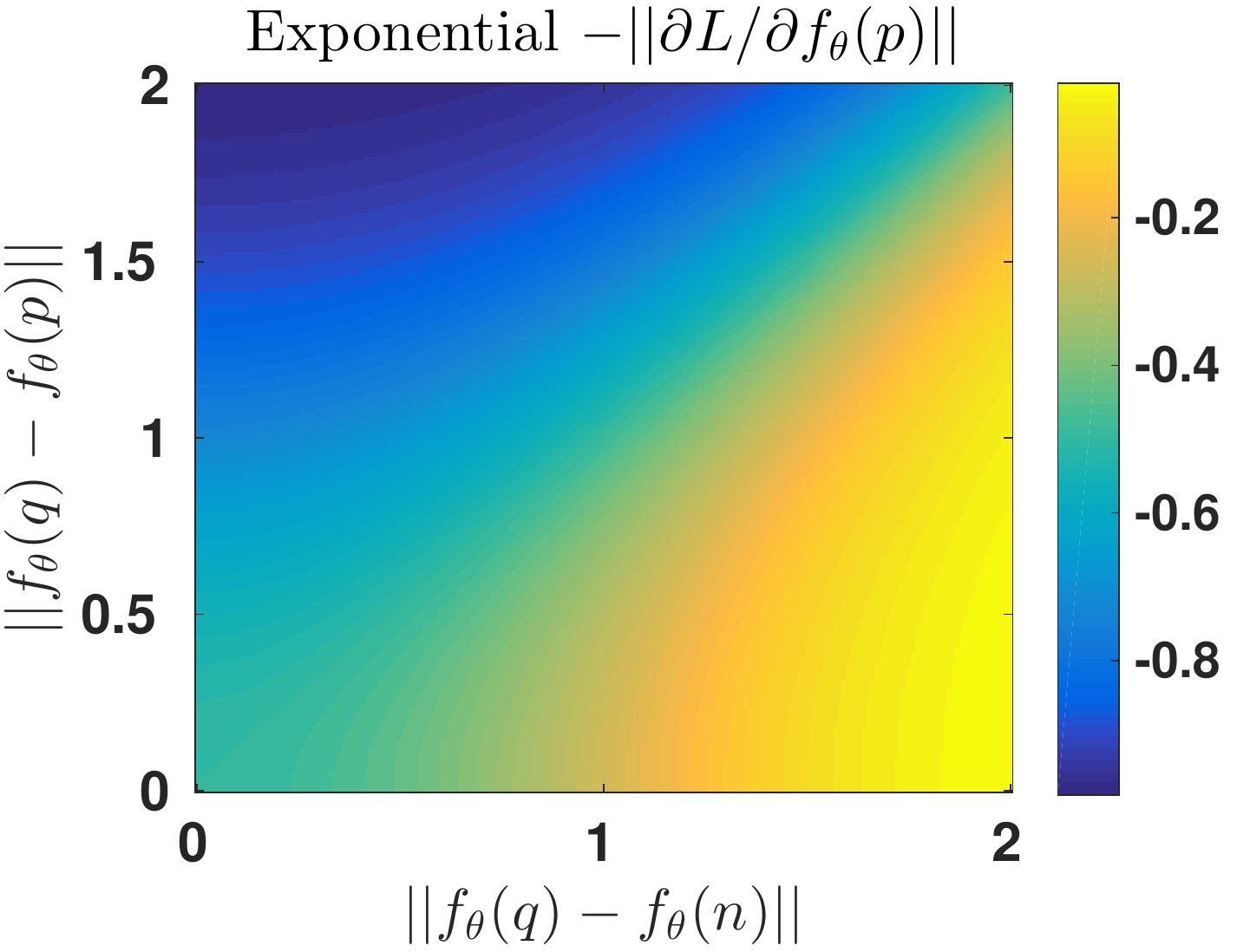}
\includegraphics[width=0.185\textwidth]{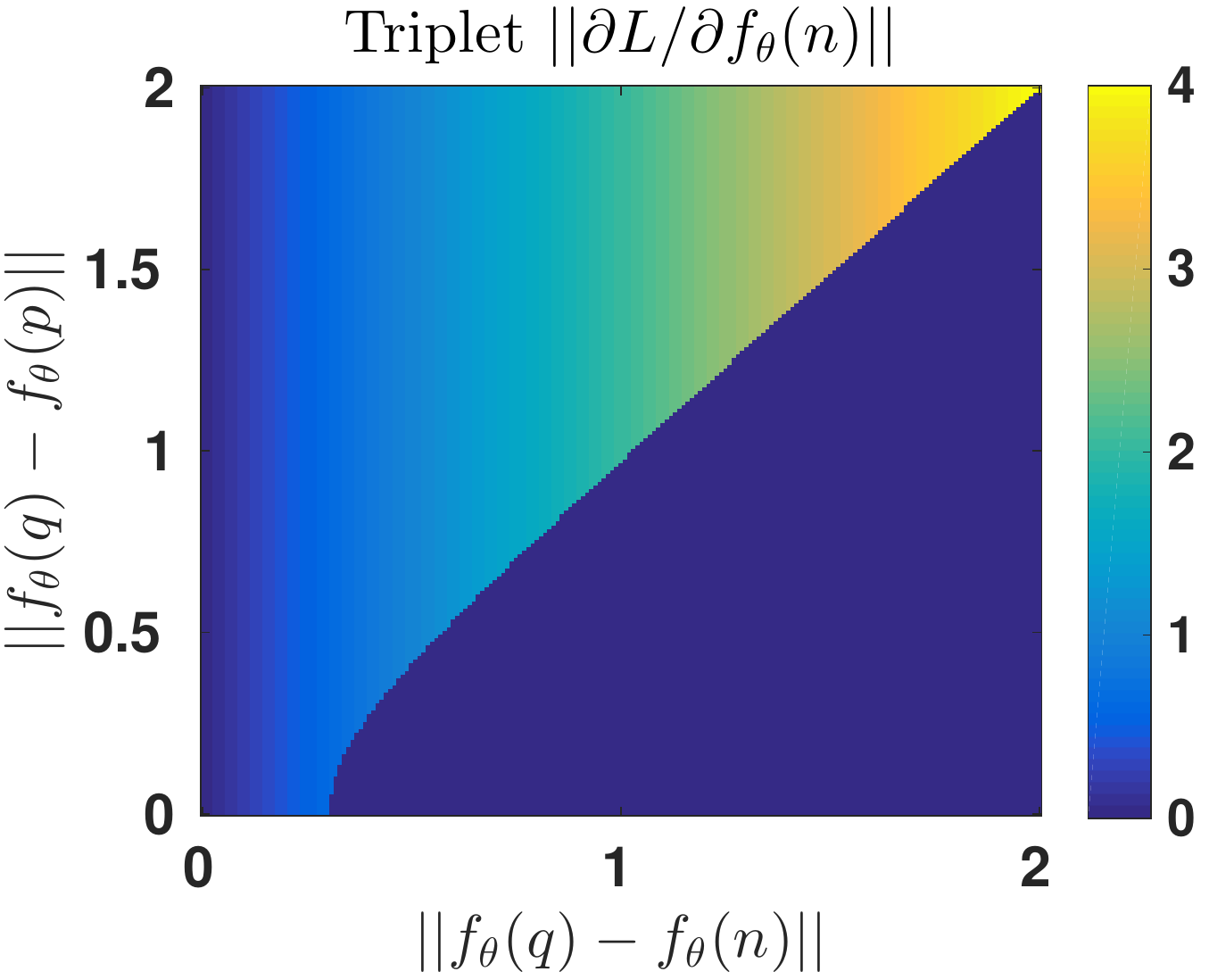}
\includegraphics[width=0.195\textwidth]{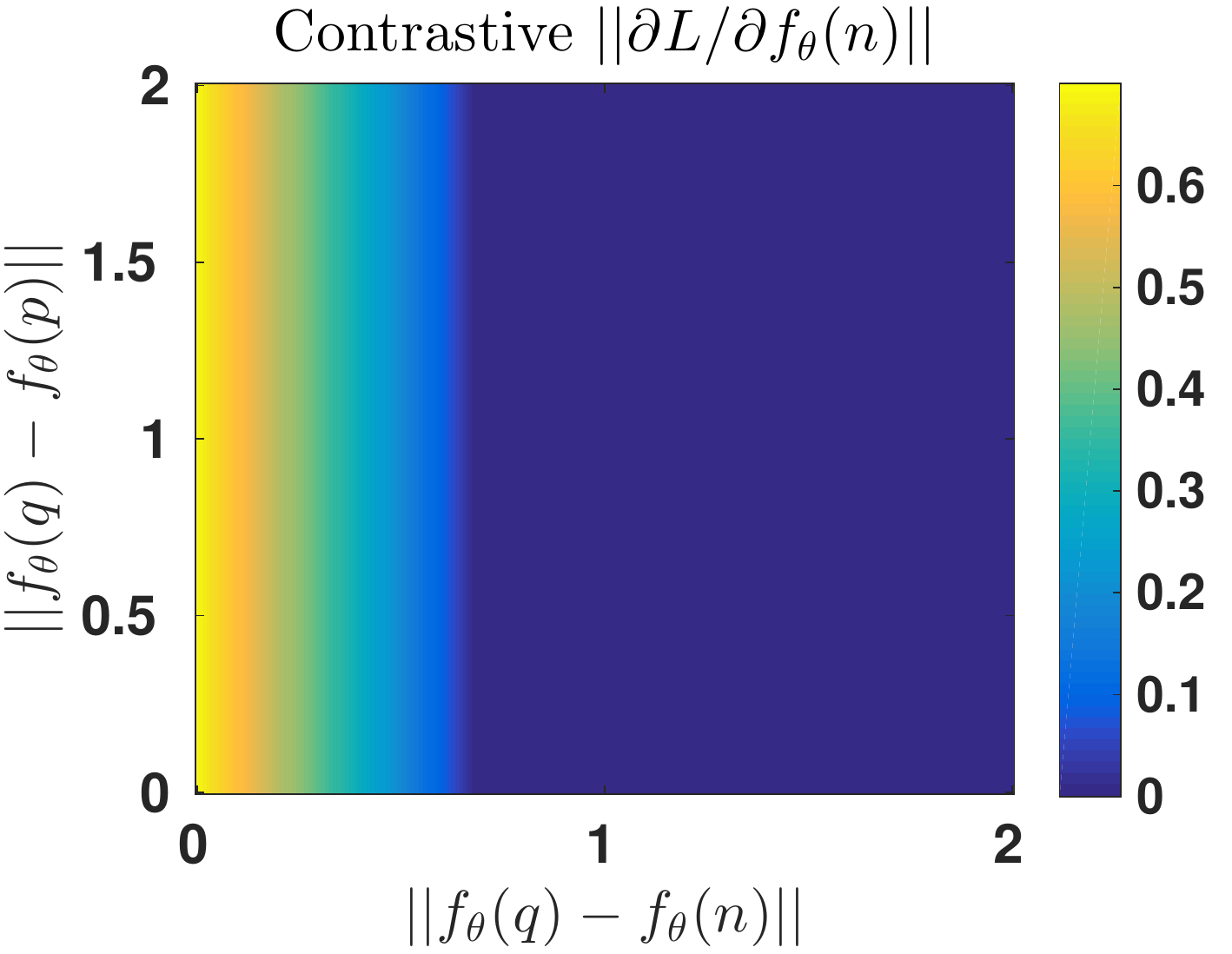}
\includegraphics[width=0.195\textwidth]{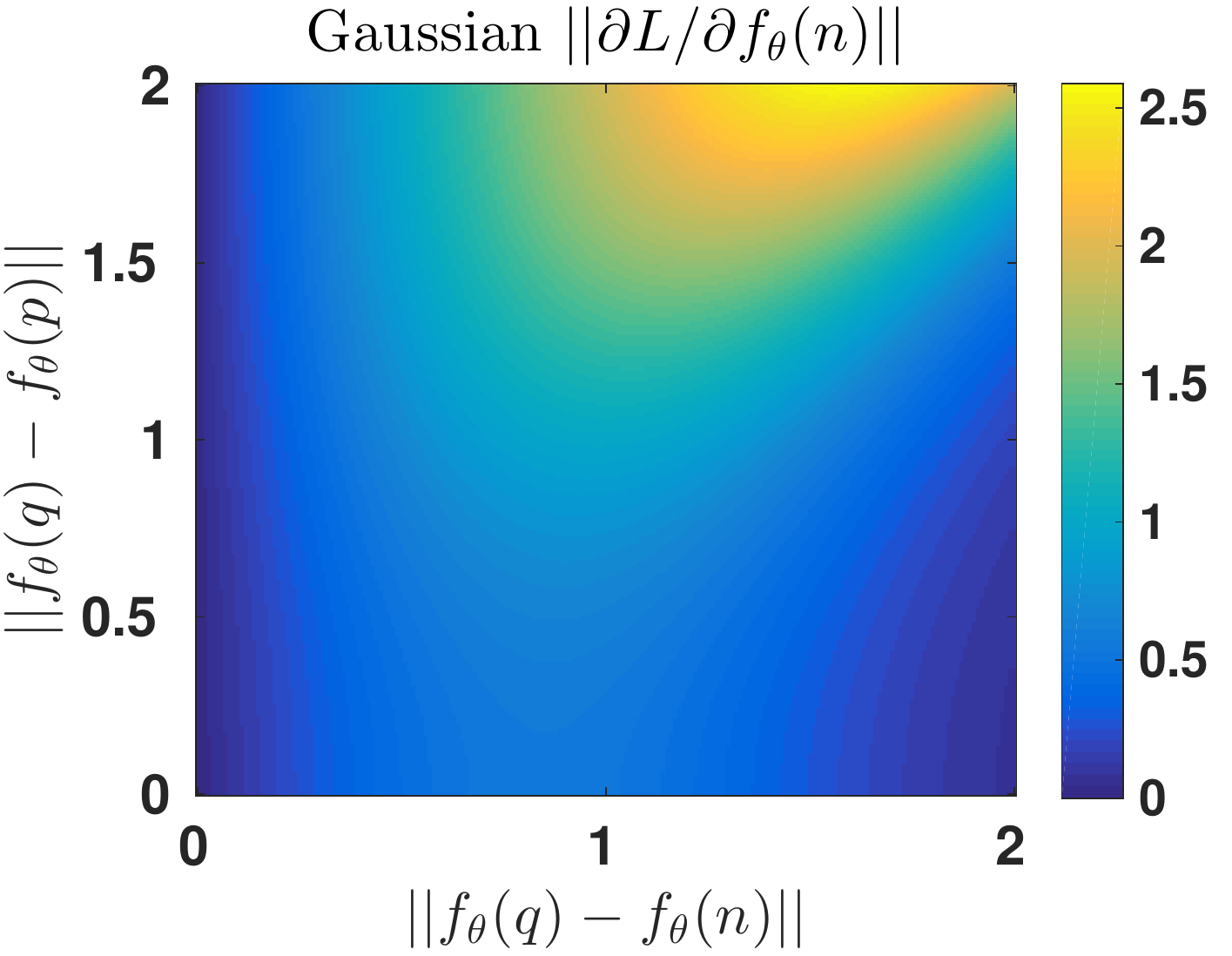}
\includegraphics[width=0.195\textwidth]{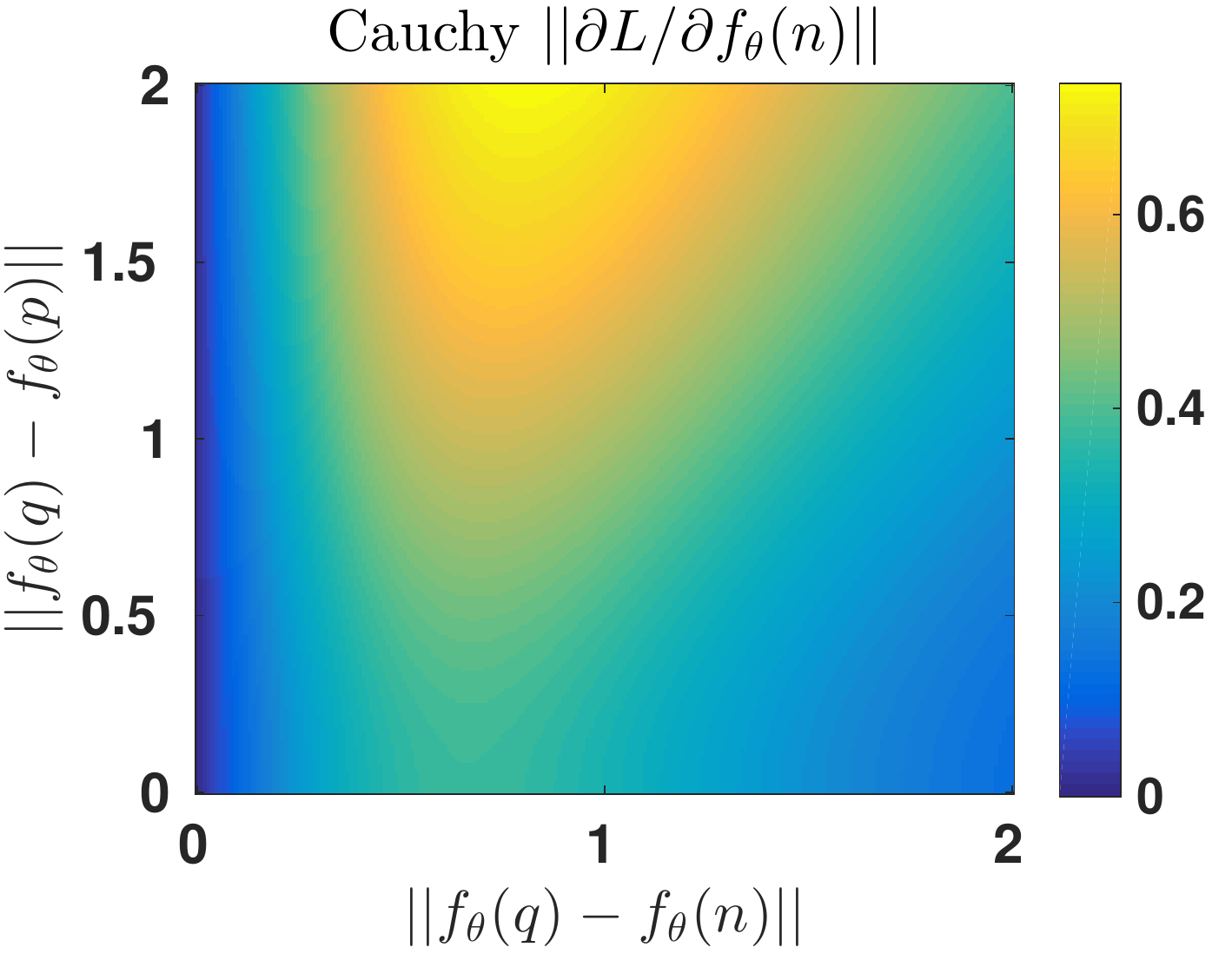}
\includegraphics[width=0.195\textwidth]{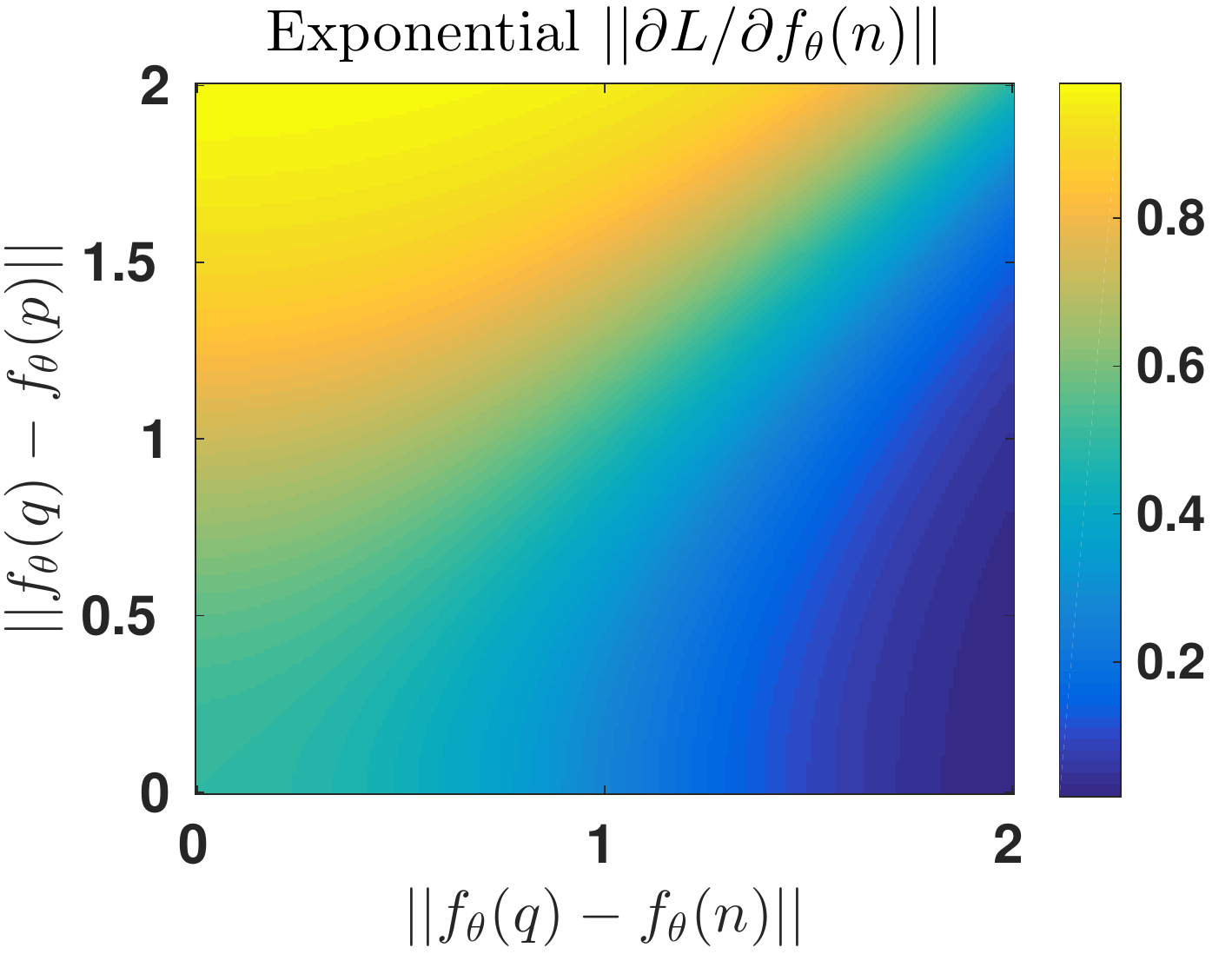}
\end{center}
\caption{Comparison of gradients with respect to $p$ and $n$ for different objectives. $m = 0.1, \tau = 0.7$. (Best viewed in color on screen)
}
\label{fig:gradients}
\end{figure*}

In Fig.~\ref{fig:gradients}, we compare the magnitudes of gradients with respect to $p$ and $n$ for different objectives. The mathematical equations of gradients with respect to $p$ and $n$ for different objectives are given in Table \ref{tab::gradients}. For each objective, the gradient with respect to $q$ is given by ${\partial L}/{\partial f_\theta(q)} =- {\partial L}/{\partial f_\theta(p)} - {\partial L}/{\partial f_\theta(n)}$.

\begin{table*}[]
\renewcommand{\arraystretch}{1.5}
\scriptsize
\centering
\caption{Comparison of gradients with respect to $p$ and $n$ for different objectives. Note that $\hat{c}_{p|q}$ and $\bar{c}_{p|q}$ are different from ${c}_{p|q}$ since they are defined by Cauchy and Exponential kernels, respectively. $\hat{c}_{p|q}$ and $\bar{c}_{p|q}$ share similar form as
${c}_{p|q}$.} 
\label{tab::gradients}
\begin{tabular}{|l|c|c|}
\hline
{\backslashbox{Loss}{Gradients}} & {$ {\partial L} /\partial f_\theta(p) $} &   {${\partial L} /\partial f_\theta(n)$} \\ \hline
Triplet ranking & $2\left(f_\theta(p)-f_\theta(q)\right)$ & $2\left(f_\theta(q)-f_\theta(n)\right)$ \\ \hline
Contrastive & $f_\theta(p)-f_\theta(q)$ & $-\left ( 1-\tau/\left \| f_\theta(q)-f_\theta(n) \right \| \right )\left ( f_\theta(q)-f_\theta(n) \right )$ \\ \hline
Gaussian SARE & $2\left ( 1-c_{p|q} \right )\left(f_\theta(p)-f_\theta(q)\right)$ & $2\left ( 1-c_{p|q} \right )\left(f_\theta(q)-f_\theta(n)\right)$ \\ \hline
Cauchy SARE & $2\left ( 1-\hat{c}_{p|q} \right )\frac{f_\theta(p)-f_\theta(q)}{ 1+\left \| f_\theta(p)- f_\theta(q)\right \|^2 }$ & $2\left ( 1-\hat{c}_{p|q} \right )\frac{f_\theta(q)-f_\theta(n)}{ 1+\left \| f_\theta(q)- f_\theta(n)\right \|^2 }$ \\ \hline
Exponential SARE & $\left ( 1-\bar{c}_{p|q} \right )\frac{f_\theta(p)-f_\theta(q)}{ \left \| f_\theta(p)- f_\theta(q)\right \| }$ & $\left ( 1-\bar{c}_{p|q} \right )\frac{f_\theta(q)-f_\theta(n)}{ \left \| f_\theta(q)- f_\theta(n)\right \| }$ \\ \hline
\end{tabular}
\end{table*}

In the case of triplet ranking loss, $\left \|{\partial L} /\partial f_\theta(p)\right \|$ and $\left \|{\partial L}/{\partial f_\theta(n)}\right \|$ increase linearly with respect to the distance $\left \| f_\theta(q)- f_\theta(p)\right \|$ and  $\left \| f_\theta(q)- f_\theta(n)\right \|$, respectively.
The saturation regions in which gradients equal to zero correspond to triplet images producing a zero loss (Eq.~\eqref{eq::triplet_violating}). 
For triplet images producing a non-zero loss, $\left \|{\partial L} /\partial f_\theta(p)\right \|$ is independent of $n$, and 
vice versa.
Thus, the updating of $f_\theta(p)$ disregards the current embedded position of $n$ and vice versa.

For the contrastive loss, $\left \|{\partial L} /\partial f_\theta(p)\right \|$ is independent of $n$ and increase linearly with respect to distance $\left \| f_\theta(q)- f_\theta(p)\right \|$ 
. $\left \|{\partial L}/{\partial f_\theta(n)}\right \|$ decreases linearly with respect to distance $\left \| f_\theta(q)- f_\theta(n)\right \|$ 
. The area in which $\left \|{\partial L}/{\partial f_\theta(n)}\right \|$ equals zero corresponds to negative images with $\left \| f_\theta(q)-f_\theta(n) \right \| > \tau$.

For all kernel defined SAREs, $\left \|{\partial L} /\partial f_\theta(p)\right \|$ and $\left \|{\partial L} /\partial f_\theta(n)\right \|$ depend on distances $\left \| f_\theta(q)- f_\theta(p)\right \|$ and $\left \| f_\theta(q)- f_\theta(n)\right \|$. The implicitly respecting of the distances comes from the probability $c_{p|q}$ (Eq.~\eqref{Eq:Gaussian:cpq}). Thus, the updating of $f_\theta(p)$ and $f_\theta(n)$ considers the current embedded positions of triplet images, which is beneficial for the possibly diverse feature distribution in the embedding space. 

The benefit of kernel defined SARE-objectives can be better understood when combined with hard-negative mining strategy, which is widely used in CNN training. The strategy returns a set of hard negative images (\ie nearest negatives in $L_2$-metric) for training. Note that both the triplet ranking loss and contrastive loss rely on empirical parameters ($m,\tau$) to prune out negatives (\cf the saturation regions). In contrast, our kernel defined SARE-objectives do not rely on these parameters. They preemptively consider the current embedded positions. For example, hard negative with $\left \| f_\theta(q)- f_\theta(p)\right \| > \left \| f_\theta(q)- f_\theta(n)\right \|$ (top-left-triangle in gradients figure) will trigger large force to pull $q\sim p$ pair while pushing $q\sim n$ pair. ``semi-hard'' \cite{schroff2015facenet} negative with $\left \| f_\theta(q)- f_\theta(p)\right \| < \left \| f_\theta(q)- f_\theta(n)\right \|$ (bottom-right-triangle in gradients figure) will still trigger force to pull $q\sim p$ pair while pushing $q\sim n$ pair, however, the force decays with increasing $\left \| f_\theta(q)- f_\theta(n)\right \|$. Here, large $\left \| f_\theta(q)- f_\theta(n)\right \|$ may correspond to well-trained samples or noise, and the gradients decay ability has the potential benefit of reducing over-fitting.

\begin{figure}
\centering
\includegraphics[width=0.35\textwidth]{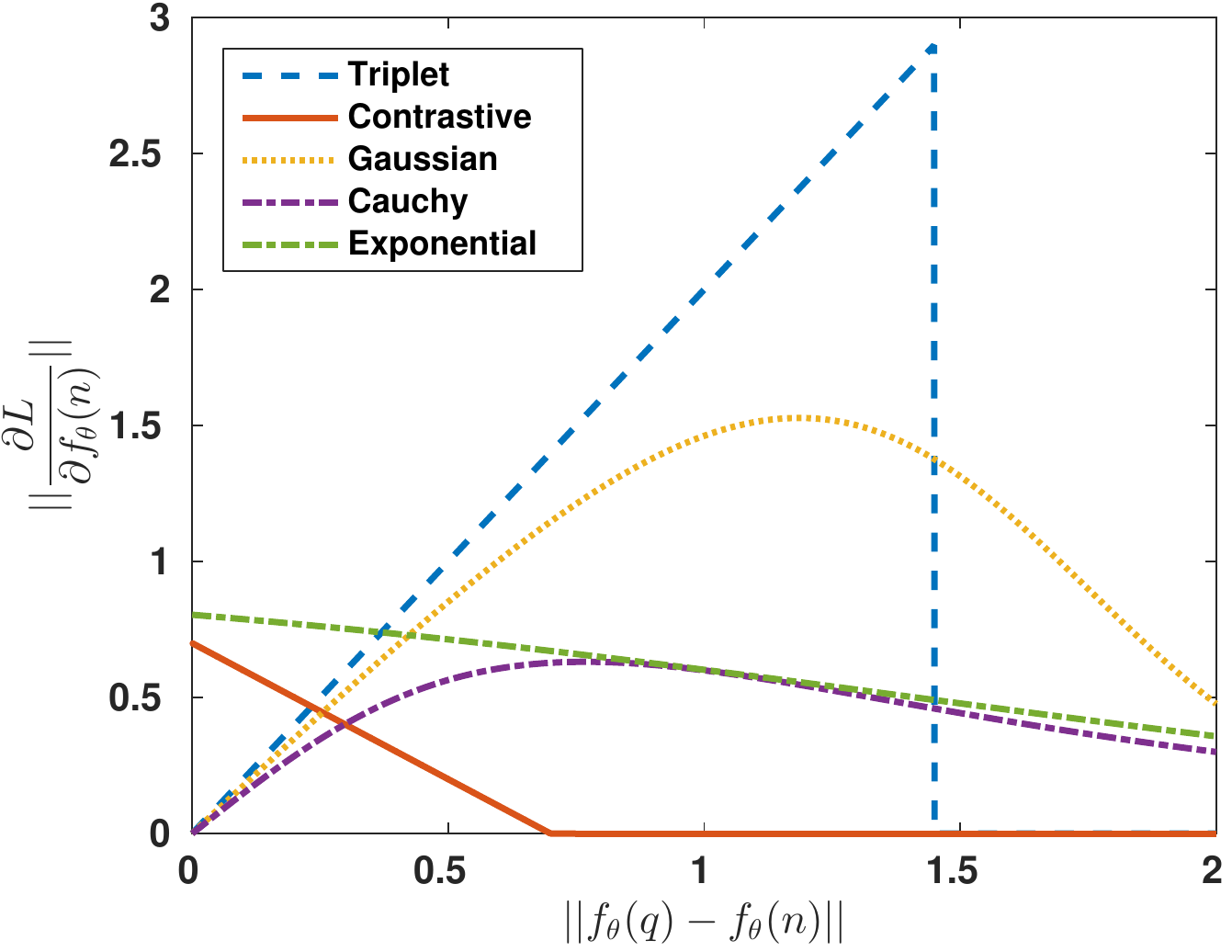}
\caption{{ Comparison of the gradients with respect to $n$ for different objectives. $m = 0.1, \tau = 0.7$. }}
\label{fig:gradients_compare}
\end{figure}

To better understand the gradient decay ability of kernel defined SARE objectives, we fix $\left \| f_\theta(q)- f_\theta(p)\right \| = \sqrt[]{2}$, and compare $\left \|{\partial L} /\partial f_\theta(n)\right \|$ for all objectives in Fig.~\ref{fig:gradients_compare}. Here, $\left \| f_\theta(q)- f_\theta(p)\right \| = \sqrt[]{2}$ means that for uniformly distributed feature embeddings, if we randomly sample $q\sim p$ pair, we are likely to obtain samples that are $\sqrt[]{2}$-away \cite{manmatha2017sampling}. Uniformly distributed feature embeddings correspond to an initial untrained/un-fine-tuned CNN.
For triplet ranking loss, Gaussian SARE and Cauchy SARE, $\left \|{\partial L} /\partial f_\theta(n)\right \|$ increases with respect to $\left \| f_\theta(q)- f_\theta(n)\right \|$ when it is small. In contrast to the gradually decay ability of SAREs, triplet ranking loss suddenly ``close'' the force when the triplet images produce a zero loss (Eq.~\eqref{eq::triplet_violating}).
For contrastive loss and Exponential SARE, $\left \|{\partial L} /\partial f_\theta(n)\right \|$ decreases with respect to $\left \| f_\theta(q)- f_\theta(n)\right \|$. Again, the contrastive loss ``close'' the force when the negative image produces a zero loss.

\section{Handling Multiple Negatives} \label{sec::SNE_extension}
\begin{figure}
\centering
\includegraphics[width=0.35\textwidth]{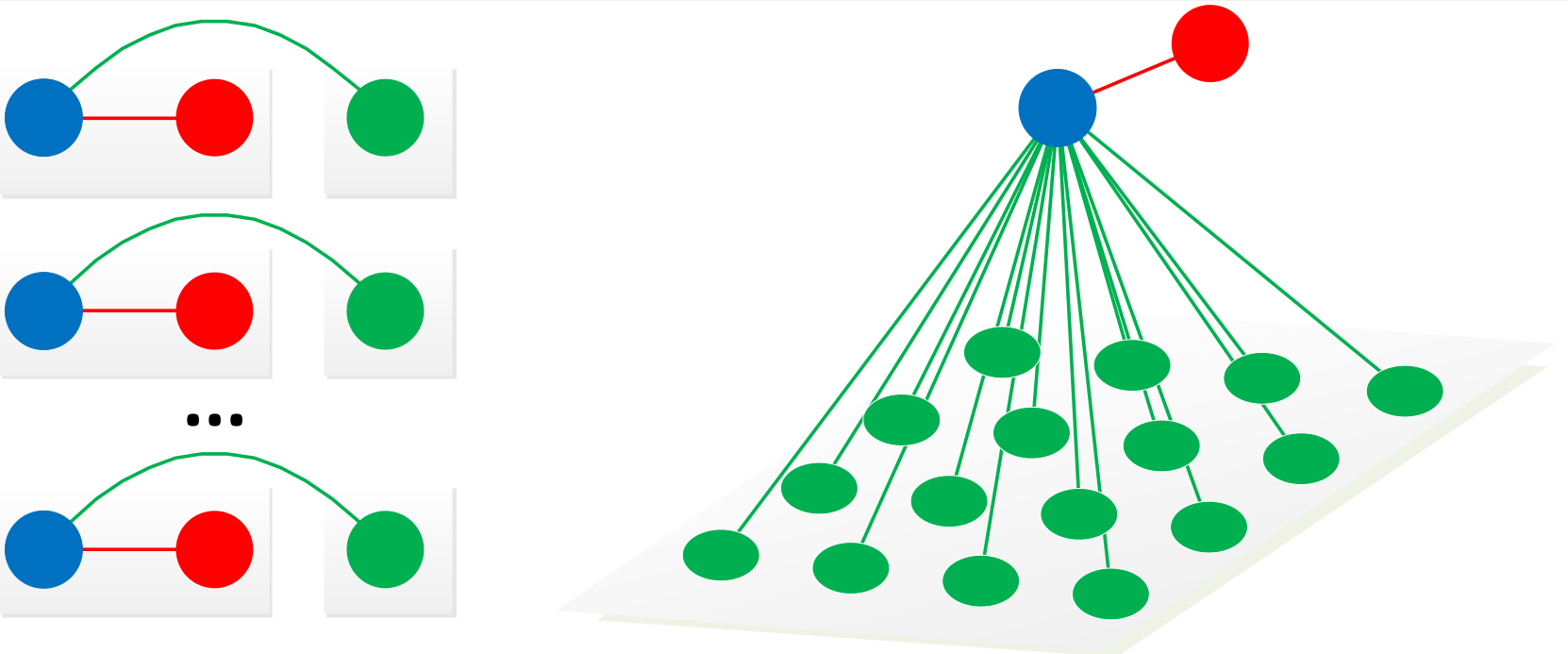}
\caption{\small {Handling multiple negative images. \textbf{Left}: The first method treats multiple negatives independently. Each triplet focuses on the competitiveness over two places, one defined by query {\Large $ \textcolor{blue}\bullet$} and positive {\Large $ \textcolor{red}\bullet$}, and the other one defined by negative {\Large $ \textcolor{green}\bullet$}. \textbf{Right}: The second strategy jointly handles multiple negative images, which enables competitiveness over multiple places.}}
\label{fig:multi_negs}
\end{figure}

In this section, we give two methods to handle multiple negative images in CNN training stage. 
Equation \eqref{triplet_loss} defines a SARE loss on a triplet and aims to shorten the embedded distance between the query and positive images while enlarging the distance between the query and negative images. Usually, in the task of IBL, the number of positive images is very small since they should depict same landmarks as the query image while the number of negative images is very big since images from different places are negative. At the same time, the time-consuming hard negative images mining process returns multiple negative images for each query image \cite{arandjelovic2016netvlad,kim2017crn}. There are two ways to handle these negative images: one is to treat them independently and the other is to jointly handle them, where both strategies are illustrated in Fig.~\ref{fig:multi_negs}.

Given $N$ negative images, treating them independently results in $N$ triplets, and they are substituted to Eq.~\eqref{triplet_loss} to calculate the loss to train CNN. Each triplet focuses on the competitiveness of two places (positive \textit{VS} negative). The repulsion and attractive forces from multiple place pairs are averaged to balance the embeddings.

Jointly handling multiple negatives aims to balance the distance of positives over multiple negatives. In our formulation, we can easily construct an objective function to push $N$ negative images simultaneously. Specifically, the match-ability priors for all the negative images are defined as zero, \ie $h_{n|q} = 0, n = 1,2,...,N$. The Kullback-Leibler divergence loss over multiple negatives is given by:
\begin{equation} \label{multiNegative_loss0}
L_\theta\left (q,p,n  \right )  =-\log\left ( c^{\ast}_{p|q} \right ),
\end{equation}
where for Gaussian kernel SARE, $c^{\ast}_{p|q}$ is defined as:
\begin{equation}
\scriptsize
c^{\ast}_{p|q} = \frac{\exp\left ( -\left \| f_\theta(q)- f_\theta(p)\right \|^2 \right )}{\exp\left ( -\left \| f_\theta(q)- f_\theta(p)\right \|^2 \right )+ \sum_{n=1}^{N}\exp\left ( -\left \| f_\theta(q)- f_\theta(n)\right \|^2 \right )}. 
\end{equation}
The gradients of Eq.~\eqref{multiNegative_loss0} can be easily computed to train CNN. 

\section{Experiments}\label{sec::Experiments}

This section mainly discusses the performance of SARE objectives for training CNN. We show that with SARE, we can improve the IBL performance on various standard place recognition and image retrieval datasets.

\subsection{Implementation Details}\label{sec::implentation}

\paragraph{Datasets.}
Google Street View Time Machine datasets have been widely-used in IBL \cite{torii201524,arandjelovic2016netvlad,kim2017crn}. It provides multiple street-level panoramic images taken at different times at close-by spatial locations on the map. The panoramic images are projected into multiple perspective images, yielding the training and testing datasets. Each image is associated with a GPS-tag giving its approximate geographic location, which can be used to identify nearby images not necessarily depicting the same landmark. We follow \cite{arandjelovic2016netvlad,Vo_2017_ICCV} to identify the positive and negative images for each query image. For each query image, the positive image is the closest neighbor in the feature embedding space at its nearby geo-position, and the negatives are far away images. The above positive-negative mining method is very efficient despite some outliers may exist in the resultant positive/negative images. If accurate positives and negatives are needed, pairwise image matching with geometric validation \cite{kim2017crn} or SfM reconstruction \cite{radenovic2016cnn} can be used. However, they are time-consuming.

The Pitts30k-training dataset \cite{arandjelovic2016netvlad} is used to train CNN, which has been shown to obtain best CNN \cite{arandjelovic2016netvlad}. To test our method for IBL, the Pitts250k-test \cite{arandjelovic2016netvlad}, TokyoTM-val \cite{arandjelovic2016netvlad}, 24/7 Tokyo \cite{torii201524} and Sf-0 \cite{chen2011city,sattler2017large} datasets are used. To show the generalization ability of our method for image retrieval, the Oxford 5k \cite{philbin2007object}, Paris 6k \cite{philbin2008lost}, and Holidays \cite{jegou2008hamming} datasets are used.

\paragraph{CNN Architecture.}
We use the widely-used compact feature vector extraction method NetVLAD \cite{arandjelovic2016netvlad,Noh_2017_ICCV,kim2017crn,sattler2017large,sattler2017benchmarking} to demonstrate the effectiveness of our method. 
Our CNN architecture is given in Fig.~\ref{fig:pipeline}.

\paragraph{Evaluation Metric.} For the place recognition datasets Pitts250k-test \cite{arandjelovic2016netvlad}, TokyoTM-val \cite{arandjelovic2016netvlad}, 24/7 Tokyo \cite{torii201524} and Sf-0 \cite{chen2011city}, we use the Precision-Recall curve to evaluate the performance. Specifically, for Pitts250k-test \cite{arandjelovic2016netvlad}, TokyoTM-val \cite{arandjelovic2016netvlad}, and 24/7 Tokyo \cite{torii201524}, the query image is deemed correctly localized if at least one of the top $N$ retrieved database images is within $d = 25$ meters from the ground truth position of the query image. The percentage of correctly recognized queries (Recall) is then plotted for different values of $N$. For the large-scale Sf-0 \cite{chen2011city} dataset, the query image is deemed correctly localized if at least one of the top $N$
retrieved database images shares the same building IDs ( manually labeled by \cite{chen2011city} ). For the image-retrieval datasets Oxford 5k \cite{philbin2007object}, Paris 6k \cite{philbin2008lost}, and Holidays \cite{jegou2008hamming}, the mean-Average-Precision (mAP) is reported.  

\paragraph{Training Details.} We use the training method of \cite{arandjelovic2016netvlad} to compare different objectives. For the state-of-the-art triplet ranking loss, the off-the-shelf implementation \cite{arandjelovic2016netvlad} is used. For the contrastive loss \cite{radenovic2016cnn}, triplet images are partitioned into $q \sim p$ and $q \sim n$ pairs to calculate the loss (Eq.~\eqref{Eq:ContrastiveLoss}) and gradients. For our method which treats multiple negatives independent (\textit{Our-Ind.}), we first calculate the probability $c_{p|q}$ (Eq.~\eqref{Eq:Gaussian:cpq}). $c_{p|q}$ is then used to calculate the gradients (Table \ref{tab::gradients}) with respect to the images. The gradients are back-propagated to train CNN. For our method which jointly handles multiple negatives (\textit{Our-Joint}), we use Eq.\eqref{multiNegative_loss0} to train CNN.
Our implementation is based on MatConvNet \cite{vedaldi15matconvnet}.

\begin{figure}
\centering
\includegraphics[width=0.235\textwidth]{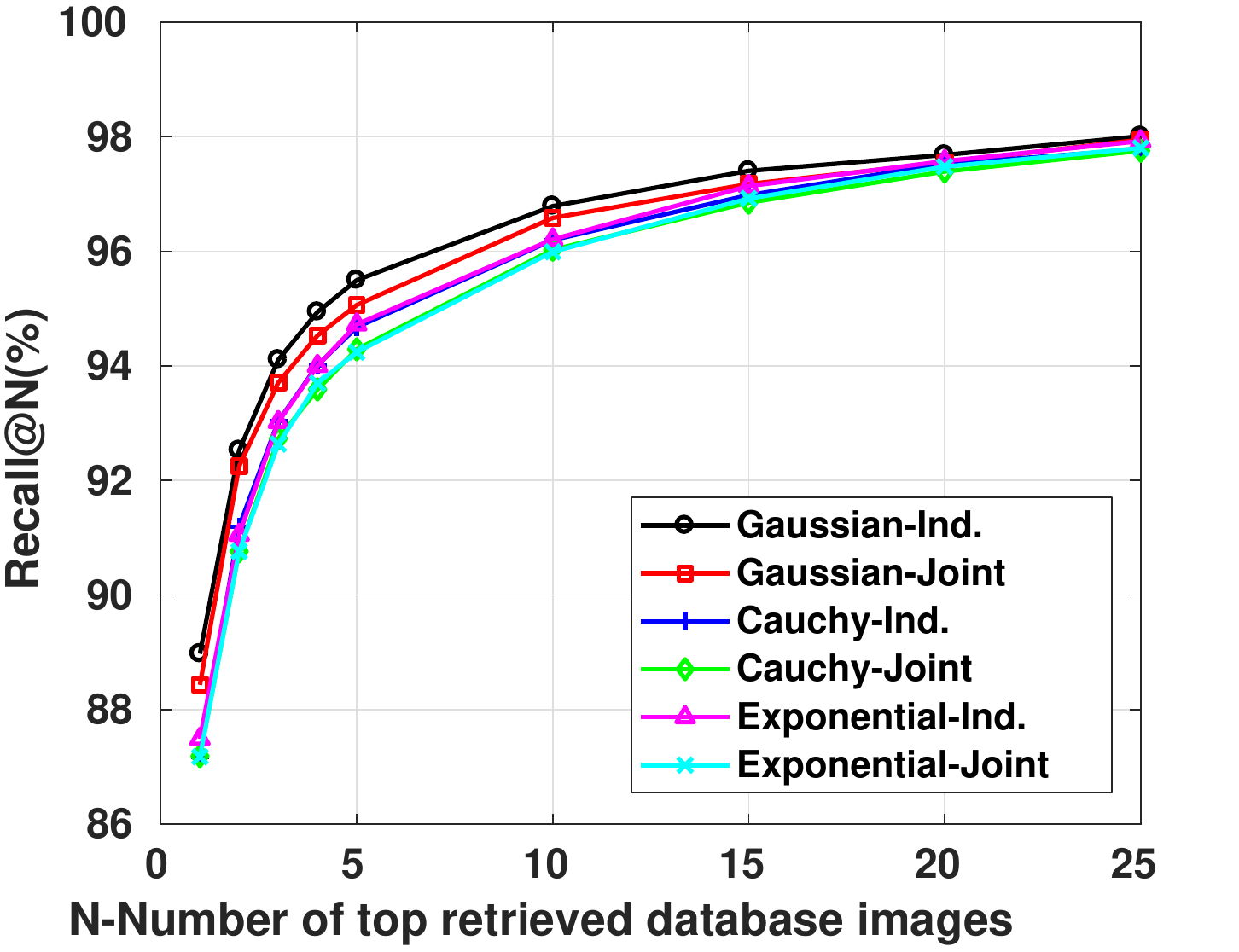}
\includegraphics[width=0.235\textwidth]{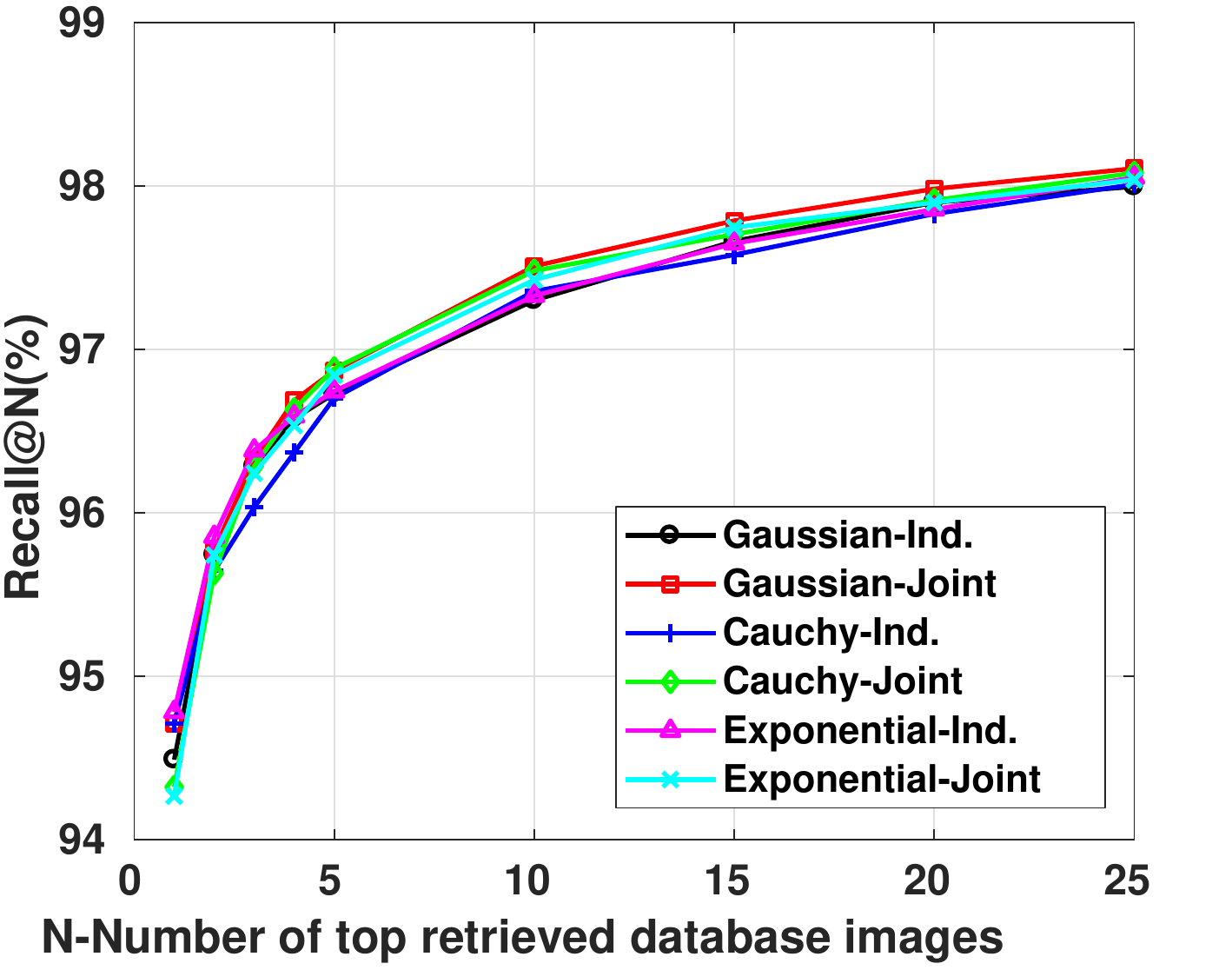}
\includegraphics[width=0.235\textwidth]{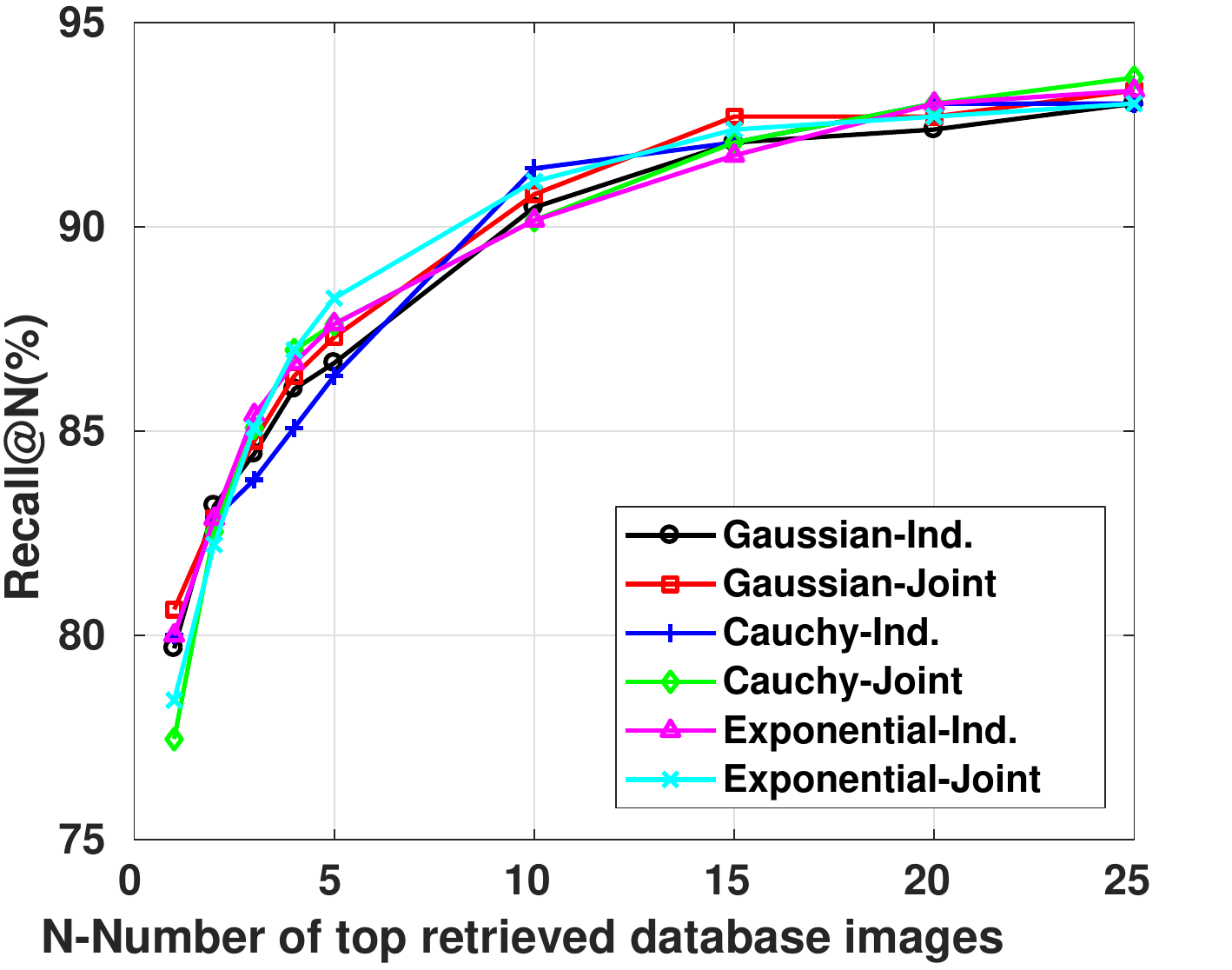}
\includegraphics[width=0.235\textwidth]{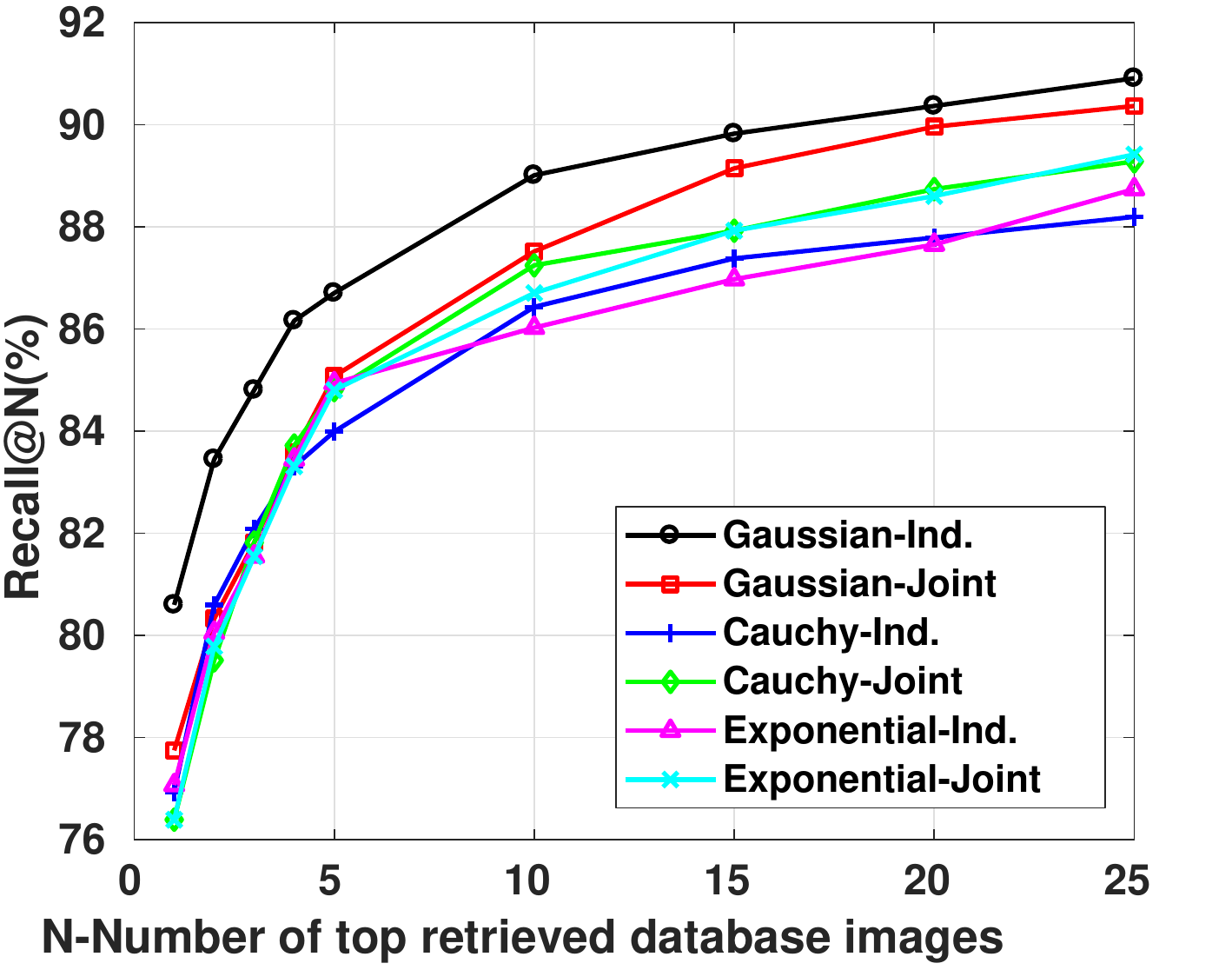}
\caption{Comparison of recalls for different kernel defined SARE-objectives. From left to right and top to down: Pitts250k-test, TokyoTM-val, 24/7 Tokyo and Sf-0. (Best viewed in color on screen) }
\label{fig:Kernels}
\end{figure}

\subsection{Kernels for SARE}
To assess the impact of kernels on fitting the pairwise $L_2$-metric feature vector distances, we compare CNNs trained by Gaussian, Cauchy and Exponential kernel defined SARE-objectives, respectively. All the hyper-parameters are the same for different objectives, and the results are given in Fig.~\ref{fig:Kernels}. CNN trained by Gaussian kernel defined SARE generally outperforms CNNs trained by others.

We find that handling multiple negatives jointly (\textit{Gaussian-Joint}) leads to better training and validation performances than handling multiple negatives independently (\textit{Gaussian-Ind.}). However, when testing the trained CNNs on Pitts250k-test, TokyoTM-val, and 24/7 Tokyo datasets, the recall performances are similar. The reason may come from the negative images sampling strategy. Since the negative images are dropped randomly from far-away places from the query image using GPS-tags, they potentially are already well-balanced in the whole dataset, thus the repulsion and attractive forces from multiple place pairs are similar, leading to a similar performance of the two methods. \textit{Gaussian-Ind.} behaves surprisingly well on the large-scale Sf-0 dataset.

\subsection{Comparison with state-of-the-art}
We use Gaussian kernel defined SARE objectives to train CNNs, and compare our method with state-of-the-art NetVLAD \cite{arandjelovic2016netvlad} and NetVLAD with Contextual Feature Reweighting \cite{kim2017crn}. The complete \textit{Recall@N} performance for different methods are given in Table \ref{Comparison_stateofart}.

\begin{table*}[]
\centering
\caption{Comparison of Recalls  on  the Pitts250k-test,  TokyoTM-val,  24/7 Tokyo and Sf-0 datasets.}
\label{Comparison_stateofart}
\begin{tabular}{|l|c|c|c|c|c|c|c|c|c|c|c|c|}
\hline
\multirow{2}{*}{\backslashbox{Method}{Dataset}} & \multicolumn{3}{c|}{Pitts250k-test} & \multicolumn{3}{c|}{TokyoTM-val} & \multicolumn{3}{c|}{24/7 Tokyo} & \multicolumn{3}{c|}{Sf-0} \\ \cline{2-13} 
                        & r@1        & r@5        & r@10      & r@1       & r@5       & r@10     & r@1       & r@5      & r@10     & r@1     & r@5    & r@10   \\ \hline
Our-Ind.                & \textbf{88.97}      & \textbf{95.50}      & \textbf{96.79}     & 94.49     & 96.73     & 97.30    & 79.68     & 86.67    & 90.48    & \textbf{80.60}   & \textbf{86.70}  & \textbf{89.01} \\ \hline
Our-Joint               & 88.43      & 95.06      & 96.58     & \textbf{94.71}     & \textbf{96.87}     & 97.51    & \textbf{80.63}     & \textbf{87.30}    & \textbf{90.79}    & 77.75   & 85.07  & 87.52  \\ \hline
CRN \cite{kim2017crn}            & 85.50      & 93.50      & 95.50     &   -        &   -        &    -      & 75.20     & 83.80    & 87.30    &  -       &   -     &   -     \\ \hline
NetVLAD \cite{arandjelovic2016netvlad}                & 85.95      & 93.20      & 95.13     &  93.85         &  96.77         &  \textbf{97.59}       & 73.33     & 82.86    & 86.03    & 75.58        &  83.31      &  85.21      \\ \hline
\end{tabular}
\end{table*}

CNNs trained  by Gaussian-SARE objectives consistently outperform state-of-the-art CNNs by  a large  margin  on  almost all  benchmarks. For  example,  on  the challenging
24/7 Tokyo dataset, \textit{our-Ind.} trained NetVLAD achieves recall@1 of 79.68\% compared to
the second-best 75.20\% obtained by CRN \cite{kim2017crn}, \ie a relative improvement in recall of  4.48\%. On the large-scale challenging Sf-0 dataset, \textit{our-Ind.} trained NetVLAD achieves recall@1 of 80.60\% compared to
the 75.58\% obtained by NetVLAD \cite{arandjelovic2016netvlad}, \ie a relative improvement in recall of 5.02\%.  Note that we do not use the Contextual Reweighting layer to capture the ``context'' within images, which has been shown to be more effective than the original NetVLAD structure \cite{kim2017crn}. Similar  improvements  can  be  observed  in other datasets.  This  confirms  the  important  premise
of this work: formulating the IBL problem in competitive learning framework, and using SARE to supervise the CNN training process can learn discriminative yet compact image representations for IBL.
We visualize 2D feature embeddings of query images from 24/7 Tokyo and Sf-0 datasets. Images taken from the same place are mostly embedded to nearby 2D positions despite the significant variations in viewpoint, pose, and configuration.

\subsection{Qualitative Evaluation}
\begin{figure*}
\centering
\begin{subfigure}[t]{\textwidth}
\centering
\begin{subfigure}[t]{0.15\textwidth}
\includegraphics[width=\textwidth]{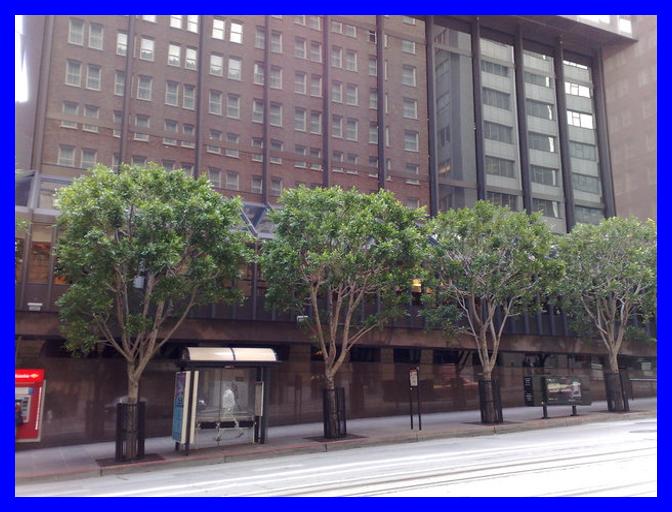}
\end{subfigure}
\begin{subfigure}[t]{0.15\textwidth}
\includegraphics[width=\textwidth]{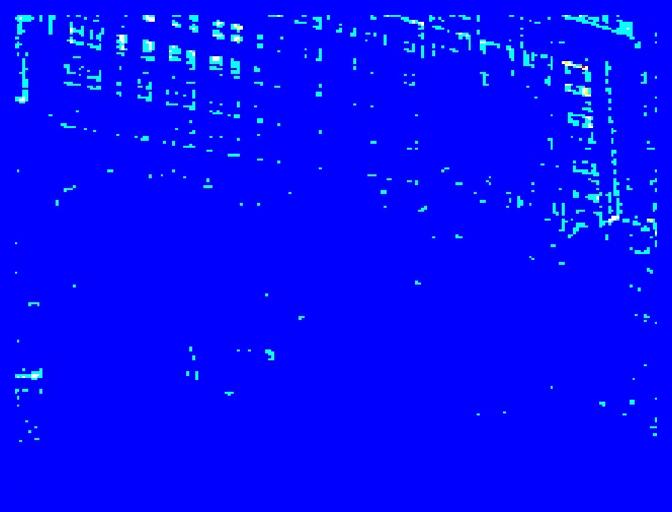}
\end{subfigure}
\begin{subfigure}[t]{0.15\textwidth}
\includegraphics[width=\textwidth]{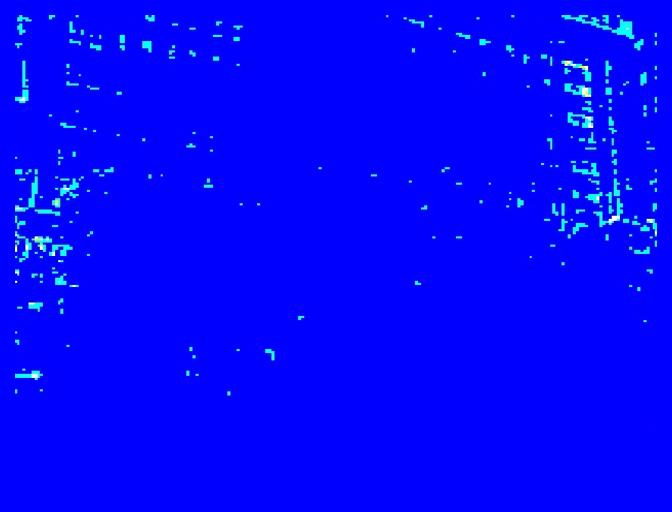}
\end{subfigure}
\begin{subfigure}[t]{0.15\textwidth}
\includegraphics[width=\textwidth]{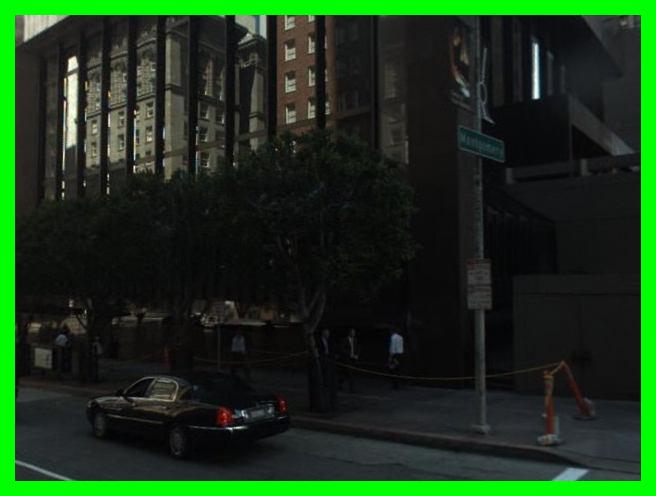}
\end{subfigure}
\begin{subfigure}[t]{0.15\textwidth}
\includegraphics[width=\textwidth]{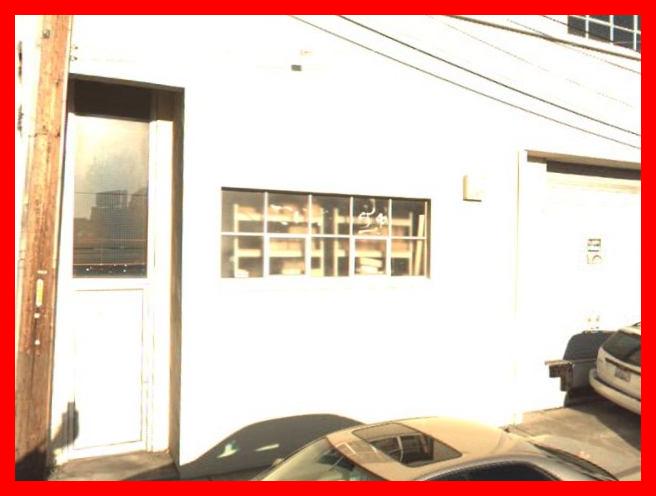}
\end{subfigure}
\end{subfigure}



\begin{subfigure}[t]{\textwidth}
\centering
\begin{subfigure}[t]{0.15\textwidth}
\includegraphics[width=\textwidth]{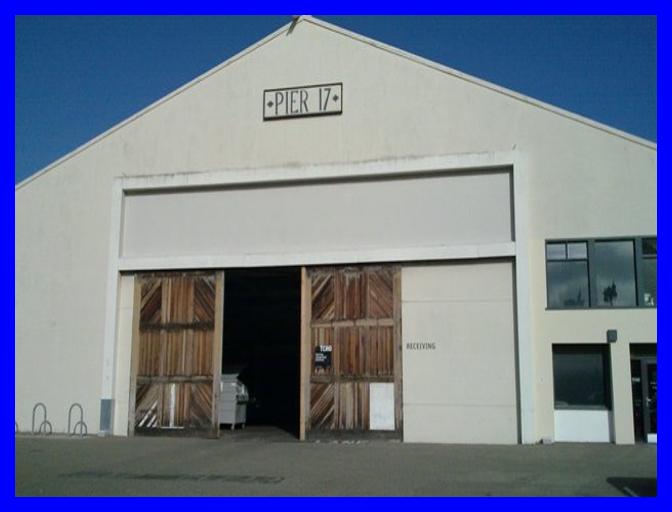}
{\caption{ Query}}
\end{subfigure}
\begin{subfigure}[t]{0.15\textwidth}
\includegraphics[width=\textwidth]{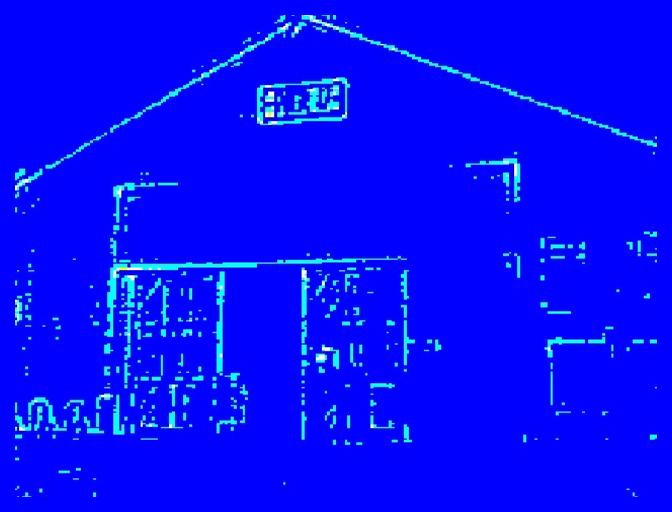}
{\caption{Our-heatmap}}
\end{subfigure}
\begin{subfigure}[t]{0.15\textwidth}
\includegraphics[width=\textwidth]{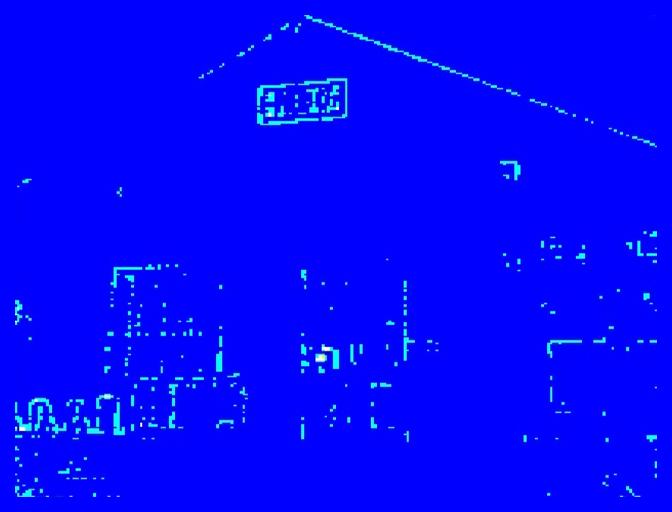}
\caption{NetVLAD-heatmap}
\end{subfigure}
\begin{subfigure}[t]{0.15\textwidth}
\includegraphics[width=\textwidth]{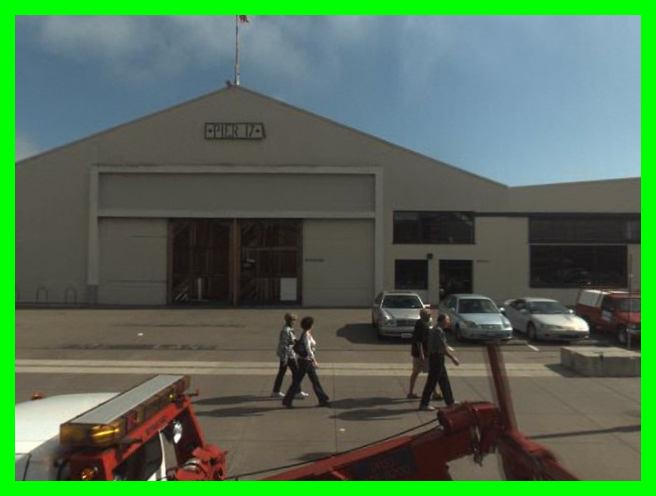}
\caption{Our-top1}
\end{subfigure}
\begin{subfigure}[t]{0.15\textwidth}
\includegraphics[width=\textwidth]{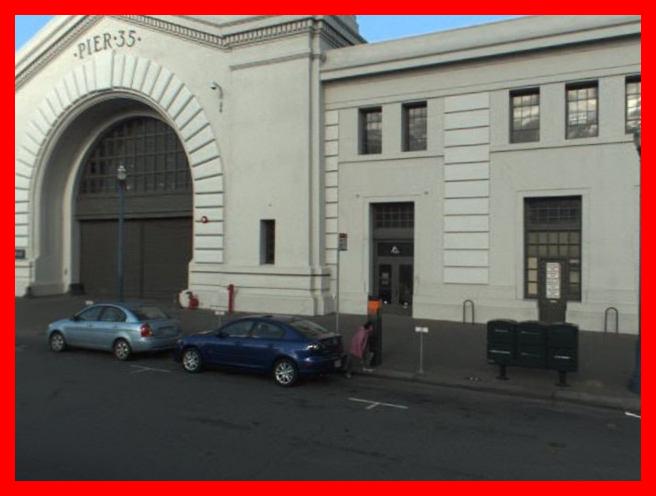}
\caption{NetVLAD-top1}
\end{subfigure}
\end{subfigure}
\caption{ Example retrieval results on Sf-0 benchmark dataset. From left to right: query image, the heat map of \textit{Our-Ind}, the heat map of NetVLAD \cite{arandjelovic2016netvlad}, the top retrieved image using our method, the top retrieved image using NetVLAD.  Green and red borders indicate correct and incorrect retrieved results, respectively. (Best viewed in color on screen) }
\label{fig:imageRetrieval}
\end{figure*}

To visualize the areas of the input image which are most important for localization, we adopt \cite{gruen16featurevis} to obtain a heat map showing the importance of different areas of the input image. The results are given in Fig.~\ref{fig:imageRetrieval}. As can be seen, our method focuses
on regions that are useful for image geo-localization while emphasizing the distinctive details on buildings. On the other hand, the NetVLAD \cite{arandjelovic2016netvlad} emphasizes local features, not the overall building style.

\subsection{Generalization on Image Retrieval Datasets}
To show the generalization ability of our method, we compare the compact image representations trained by different methods on standard image retrieval benchmarks (Oxford 5k \cite{philbin2007object}, Paris 6k \cite{philbin2008lost}, and Holidays \cite{jegou2008hamming}) without any fine-tuning. The results are given in Table \ref{tab:retrieval_netvlad}
. 
Comparing the CNN trained by our methods and the off-the-shelf NetVLAD \cite{arandjelovic2016netvlad} and CRN \cite{kim2017crn}, in most cases, the mAP of our methods outperforms theirs'. Since our CNNs are trained using a city-scale building-oriented dataset from urban areas, it lacks the ability to understand the natural landmarks ($\eg$ water, boats, cars), resulting in a performance drop in comparison with the city-scale building-oriented datasets. CNN trained by images similar to images encountered at test time can increase the retrieval performance \cite{babenko2014neural}. However, our purpose here is to demonstrate the generalization ability of SARE trained CNNs, which has been justified.


\begin{table}[]
\setlength{\tabcolsep}{4pt}
\centering
\caption{Retrieval performance of CNNs on image retrieval benchmarks. No spatial re-ranking or query expansion is performed. The accuracy is measured by the mean Average Precision (mAP).}
\label{tab:retrieval_netvlad}
\begin{tabular}{|l|c|c|c|c|c|}
\hline
\multirow{2}{*}{Method}  & \multicolumn{2}{c|}{Oxford 5K}  & \multicolumn{2}{c|}{Paris 6k}   & \multirow{2}{*}{Holidays} \\ \cline{2-5}
                                             & full           & crop           & full           & crop           &                           \\ \hline
Our-Ind.                                  & \textbf{71.66} & \textbf{75.51} & \textbf{82.03} & 81.07          & 80.71                     \\ \hline
Our-Joint                                & 70.26          & 73.33          & 81.32          & \textbf{81.39} & \textbf{84.33}            \\ \hline
NetVLAD \cite{arandjelovic2016netvlad}                              & 69.09          & 71.62          & 78.53          & 79.67          & 83.00                     \\ \hline
CRN  \cite{kim2017crn}                               & 69.20          & -              & -              & -              & -                         \\ \hline
\end{tabular}
\end{table}

\begin{figure}
\centering
\includegraphics[height = 0.2\textwidth, width=0.225\textwidth]{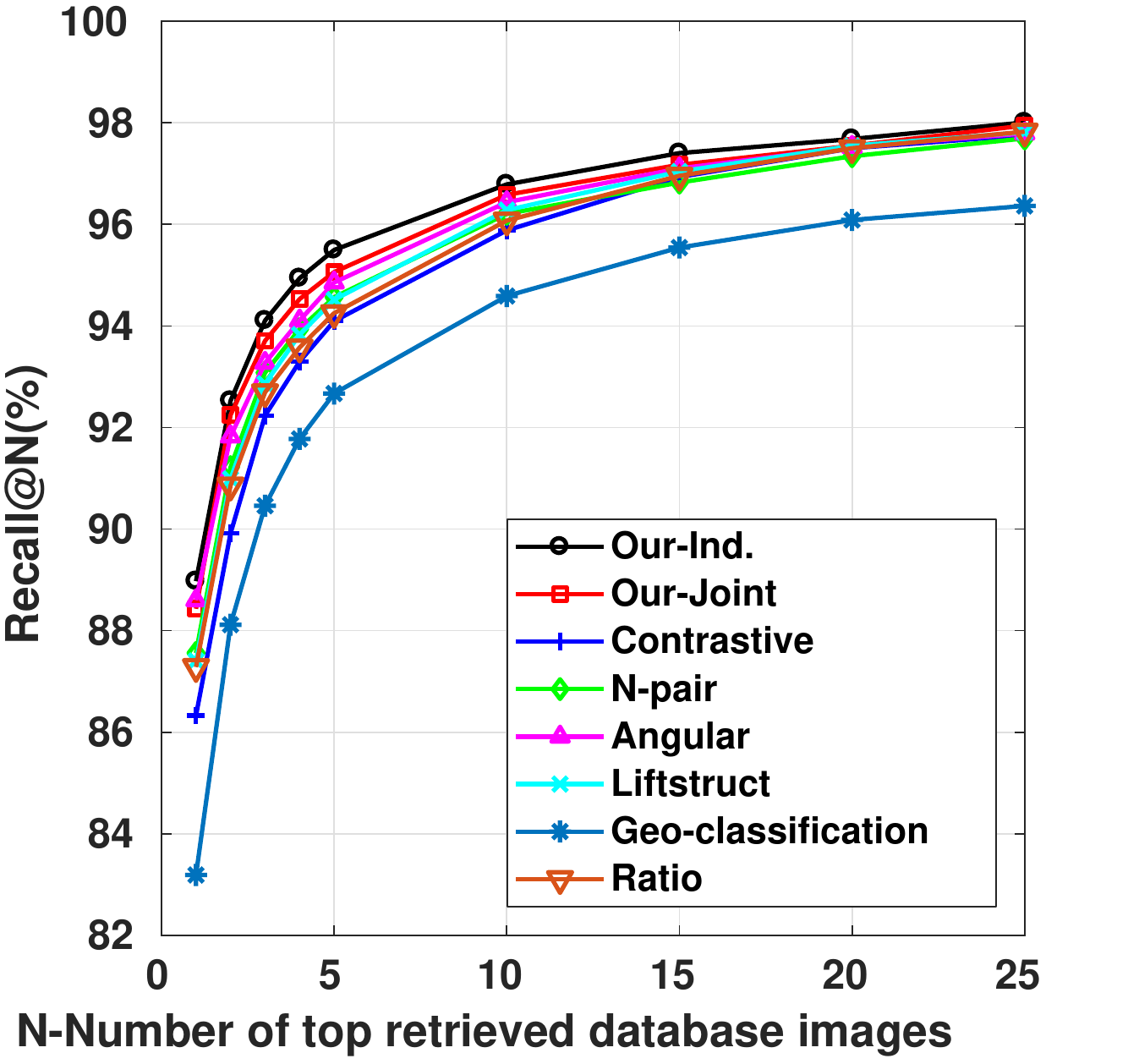}
\includegraphics[height = 0.2\textwidth, width=0.225\textwidth]{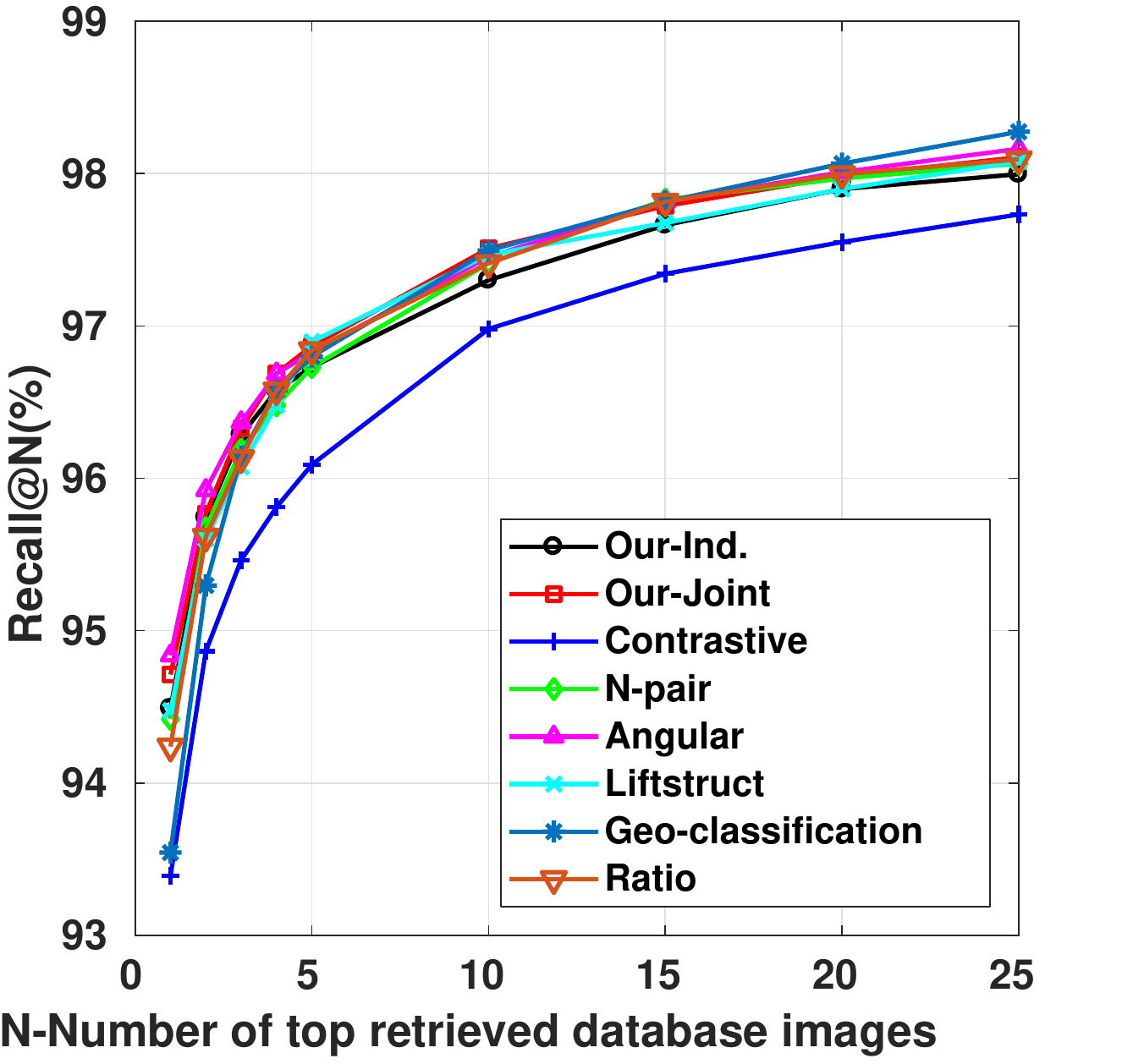}
\includegraphics[height = 0.2\textwidth, width=0.225\textwidth]{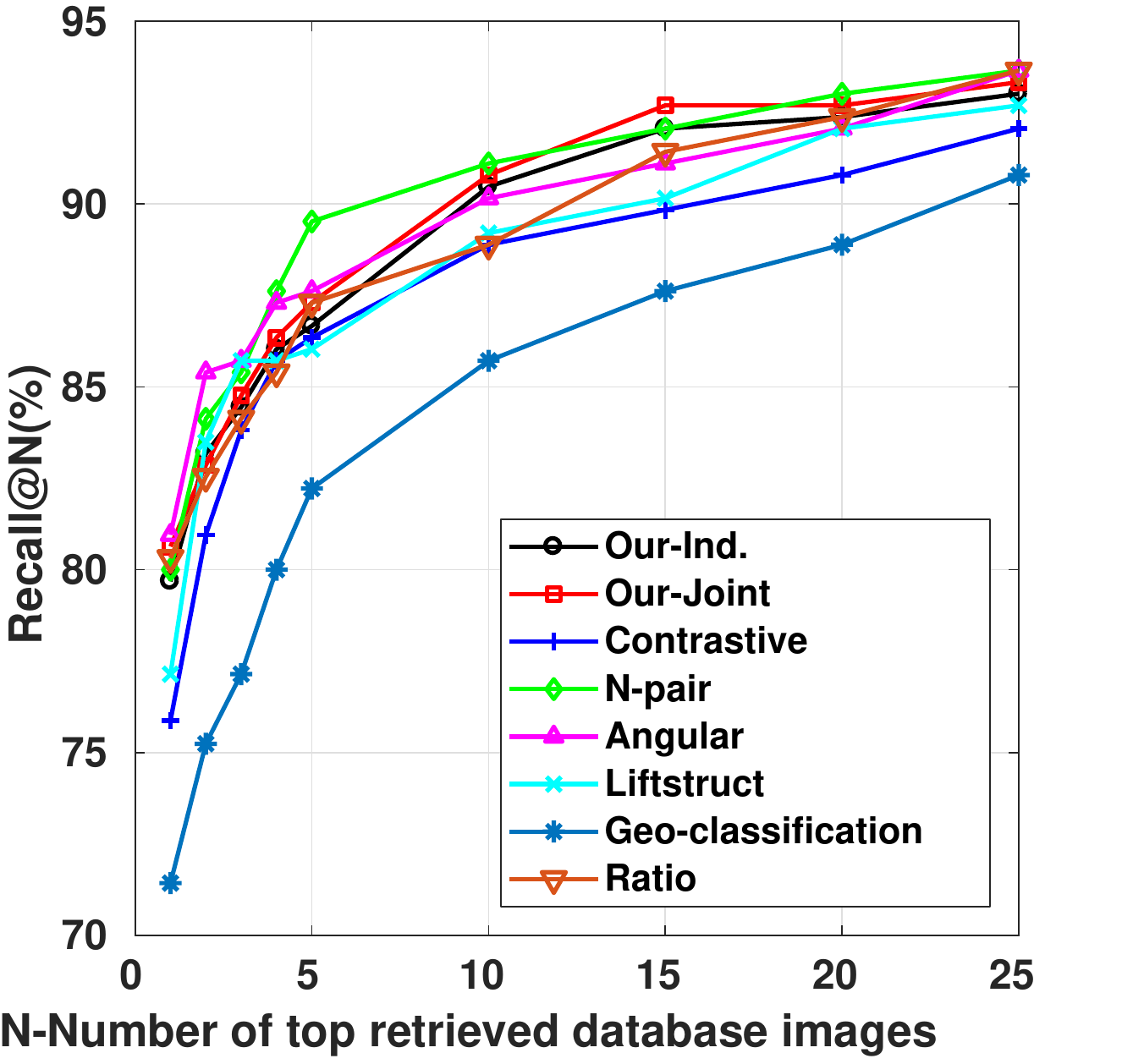}
\includegraphics[height = 0.2\textwidth, width=0.225\textwidth]{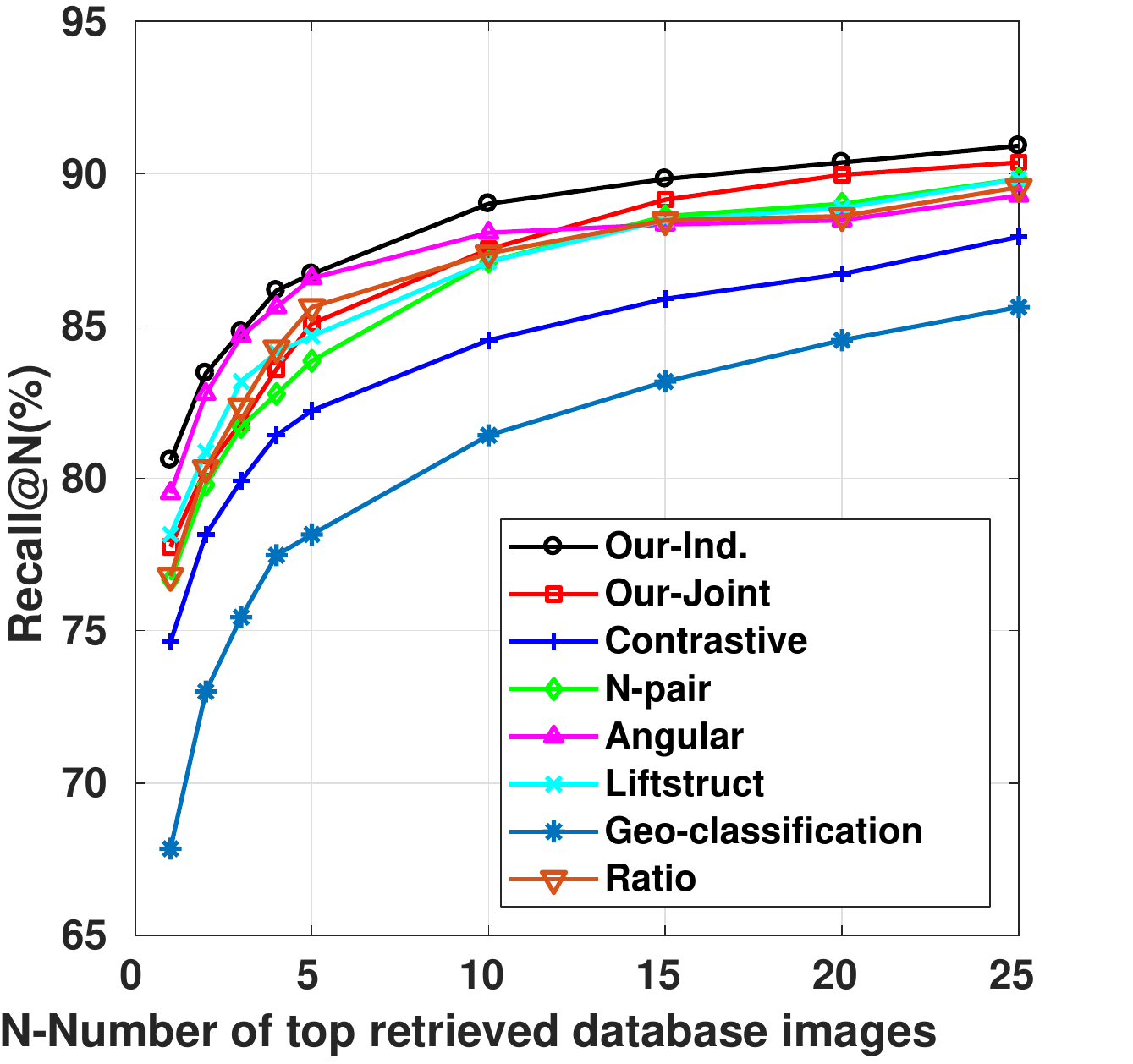}
\caption{Comparison of recalls for deep metric learning objectives. From left to right and top to down: Pitts250k-test, TokyoTM-val, 24/7 Tokyo and Sf-0.    }
\label{fig:deepmetric}
\end{figure}

\subsection{Comparison with Metric-learning Methods}

Although deep metric-learning methods have shown their effectiveness in classification and fine-grain recognition tasks, their abilities in the IBL task are unknown. As another contribution of this paper, we show the performances of six current state-of-the-art deep metric-learning methods in IBL, and compare our method with : (1) Contrastive loss used by \cite{radenovic2016cnn}; (2) Lifted structure embedding \cite{oh2016deep}; (3) N-pair loss \cite{sohn2016improved}; (4) N-pair angular loss \cite{Wang_2017_ICCV}; (5) Geo-classification loss \cite{Vo_2017_ICCV}; (6) Ratio loss \cite{hoffer2015deep}. 

Fig.~\ref{fig:deepmetric} shows the results of the quantitative comparison between our method and other deep metric learning methods.
Our theoretically-grounded method outperforms the Contrastive loss \cite{radenovic2016cnn} and Geo-classification loss \cite{Vo_2017_ICCV}, while remains comparable with other state-of-the-art methods.

\section{Conclusion}

This paper has addressed the problem of learning discriminative image representations specifically tailored for the task of Image-Based Localization (IBL). We have proposed a new Stochastic Attraction and Repulsion Embedding (SARE) objective for this task. SARE directly enforces the ``attraction" and ``repulsion" constraints on intra-place and inter-place feature embeddings, respectively. The ``attraction" and ``repulsion" constraints are formulated as a similarity-based binary classification task.  It has shown that SARE improves IBL performance, outperforming other state-of-the-art methods.

\section*{Acknowledgement}
{ This research was supported in part by the Australian Research Council (ARC) grants (CE140100016), Australia Centre for Robotic Vision, and the Natural Science Foundation of China grants
(61871325,   61420106007,   61671387,   61603303). Hongdong Li is also funded in part by ARC-DP (190102261) and ARC-LE (190100080). We
gratefully acknowledge the support of NVIDIA Corporation with the donation of the GPU. We thank all anonymous reviewers for their valuable comments.}

{\small
\bibliographystyle{ieee}
\bibliography{egbib}
}

\clearpage
\newpage

\section*{Appendix}
\appendix

In the appendix, we describe the gradients of loss functions which jointly handle multiple negative images (Sec.\ref{sec::handle_multi_negs}). Our implementation details  are given in Sec.\ref{sec::impleDetails}. Additional experimental results are given in  Sec.\ref{sec::Additional_Results}.

\section{Handling Multiple Negatives} \label{sec::handle_multi_negs}
Give a query image $q$, a positive image $p$, and multiple negative images $\{n\}, n =1,2,...,N$. The Kullback-Leibler divergence loss over multiple negatives is given by:
\begin{equation} \label{multiNegative_loss}
L_\theta\left (q,p,n  \right )  =-\log\left ( c^{\ast}_{p|q} \right ),
\end{equation}
For Gaussian kernel SARE, $c^{\ast}_{p|q}$ is defined as:

\begin{equation}\label{cpq}
\thinmuskip = 1mu
\medmuskip = 1mu
\thickmuskip = 1mu
\footnotesize
c^{\ast}_{p|q} = \frac{\exp\left ( -\left \| f_\theta(q)- f_\theta(p)\right \|^2 \right )}{\exp\left ( -\left \| f_\theta(q)- f_\theta(p)\right \|^2 \right )+ \sum_{n=1}^{N}\exp\left ( -\left \| f_\theta(q)- f_\theta(n)\right \|^2 \right )}. 
\end{equation}
where $f_\theta(q),f_\theta(p),f_\theta(n)$ are the feature embeddings of query, positive and negative images, respectively.

Substituting Eq.~\eqref{cpq} into Eq.~\eqref{multiNegative_loss} gives:
\begin{multline} \label{multiNegative_loss_gauss}
\thinmuskip = 1mu
\medmuskip = 1mu
\thickmuskip = 1mu
\footnotesize
L_\theta\left (q,p,n  \right )  = \\
 \log\left ( 1+ \sum_{n=1}^N \exp ({ \| f_\theta(q)- f_\theta(p) \|^2 - \| f_\theta(q)- f_\theta(n) \|^2 } ) \right )
\end{multline}

Denote  {\small $ 1+ \sum_{n=1}^N\exp ({\left \| f_\theta(q)- f_\theta(p)\right \|^2 -\left \| f_\theta(q)- f_\theta(n)\right \|^2 })$} as $\eta$, the gradients of Eq.~\eqref{multiNegative_loss_gauss} with respect to the query, positive and negative images are given by:

\begin{align} 
\frac{\partial L}{\partial f_\theta(p)} &=\sum_{n=1}^{N}-\frac{2}{\eta}\exp\left ({\left \| f_\theta(q)- f_\theta(p)\right \|^2 -\left \| f_\theta(q)- f_\theta(n)\right \|^2 }\right ) \nonumber \\
& \left [f_\theta(q)- f_\theta(p)\right ], \label{Eq:gauss:dldp} \\ 
\frac{\partial L}{\partial f_\theta(n)} &=\frac{2}{\eta}\exp\left ({\left \| f_\theta(q)- f_\theta(p)\right \|^2 -\left \| f_\theta(q)- f_\theta(n)\right \|^2 }\right )\nonumber \\
&\left [f_\theta(q)- f_\theta(n)\right ],\label{Eq:gauss:dldn} \\ 
\frac{\partial L}{\partial f_\theta(q)} &=- \frac{\partial L}{\partial f_\theta(p)} - \sum_{n=1}^{N}\frac{\partial L}{\partial f_\theta(n)}. \label{Eq:gauss:dldq}
\end{align}

Similarly, for Cauchy kernel, the loss function is given by:

\begin{equation} \label{multiNegative_loss_t}
L_\theta\left (q,p,n  \right )  =\log\left ( 1+ \sum_{n=1}^N\frac{1+\left \| f_\theta(q)- f_\theta(p)\right \|^2}{1+\left \| f_\theta(q)- f_\theta(n)\right \|^2} \right ).
\end{equation}

Denote  $ 1+ \sum_{n=1}^N\frac{1+ \| f_\theta(q)- f_\theta(p) \|^2}{1+ \| f_\theta(q)- f_\theta(n) \|^2} $ as $\eta$, the gradients of Eq.~\eqref{multiNegative_loss_t} with respect to the query, positive and negative images are given by:
\begin{align} 
\frac{\partial L}{\partial f_\theta(p)} &=\sum_{n=1}^{N}\frac{-2}{\eta\left ( 1+\left \| f_\theta(q)-f_\theta(n) \right \|^2 \right )} \left [f_\theta(q)-f_\theta(p)  \right ], \label{Eq:t:dldp} \\ 
\frac{\partial L}{\partial f_\theta(n)} &=\frac{2\left ( 1+ \left \| f_\theta(q)-f_\theta(p) \right \|^2 \right )}{\eta\left ( 1+\left \| f_\theta(q)-f_\theta(n) \right \|^2 \right )^2} \left [f_\theta(q)-f_\theta(n)  \right ], \label{Eq:t:dldn} \\ 
\frac{\partial L}{\partial f_\theta(q)} &=- \frac{\partial L}{\partial f_\theta(p)} - \sum_{n=1}^{N}\frac{\partial L}{\partial f_\theta(n)}. \label{Eq:t:dldq}
\end{align}

For Exponential kernel, the loss function is given by:
\begin{equation} \label{multiNegative_loss_exp}
\footnotesize
L_\theta\left (q,p,n  \right )  =\log\left ( 1+ \sum_{n=1}^N \exp\left ({\left \| f_\theta(q)- f_\theta(p)\right \| -\left \| f_\theta(q)- f_\theta(n)\right \| }\right ) \right ).
\end{equation}

Denote  $1+\sum_{n=1}^N\exp ({ \| f_\theta(q)- f_\theta(p) \| - \| f_\theta(q)- f_\theta(n)\|})$ as $\eta$, the gradients of Eq.~\eqref{multiNegative_loss_exp} with respect to the query, positive and negative images are given by:

\begin{align} 
\frac{\partial L}{\partial f_\theta(p)} &=\sum_{n=1}^{N}-\frac{\exp\left ({\left \| f_\theta(q)- f_\theta(p)\right \| -\left \| f_\theta(q)- f_\theta(n)\right \| }\right )}{\eta\left \| f_\theta(q)- f_\theta(p)\right \|} \nonumber \\ 
& \left [f_\theta(q)- f_\theta(p)\right ], \label{Eq:gauss:dldp} \\ 
\frac{\partial L}{\partial f_\theta(n)} &=\frac{\exp\left ({\left \| f_\theta(q)- f_\theta(p)\right \| -\left \| f_\theta(q)- f_\theta(n)\right \| }\right )}{\eta\left \| f_\theta(q)- f_\theta(n)\right \|} \nonumber \\ 
&\left [f_\theta(q)- f_\theta(n)\right ],
 \label{Eq:gauss:dldn} \\ 
\frac{\partial L}{\partial f_\theta(q)} &=- \frac{\partial L}{\partial f_\theta(p)} - \sum_{n=1}^{N}\frac{\partial L}{\partial f_\theta(n)}. \label{Eq:gauss:dldq}
\end{align}

The gradients are back propagated to train the CNN.

\section{Implementation Details} \label{sec::impleDetails}
We exactly follow the training method of \cite{arandjelovic2016netvlad}, without fine-tuning any hyper-parameters. The VGG-16 \cite{simonyan2014very} net is cropped at the last convolutional layer (conv5), before ReLU. 
The learning rate for the Pitts30K-train and Pitts250K-train datasets are set to 0.001 and 0.0001, respectively. They are halved every  5  epochs,  momentum  0.9,  weight  decay  0.001,  batch size of 4 tuples. Each tuple consist of one query image, one positive image, and ten negative images. The CNN is trained for at most 30 epochs but convergence usually occurs much faster (typically less than 5 epochs).  The network which yields the best recall@5 on the validation set is used for testing.

\paragraph{\textbf{Triplet ranking loss}}
For the triplet ranking loss \cite{arandjelovic2016netvlad}, we set margin $m = 0.1$, and triplet images producing a non-zero loss are used in gradient computation, which is the same as \cite{arandjelovic2016netvlad}.

\paragraph{\textbf{Contrastive loss}}
For the contrastive loss \cite{radenovic2016cnn}, we set margin $\tau = 0.7$, and negative images producing a non-zero loss are used in gradient computation. Note that positive images are always used in training since they are not pruned out.

\paragraph{\textbf{Geographic classification loss}}
For the geographic classification method \cite{Vo_2017_ICCV}, we use the Pitts250k-train dataset for training. We first partition the 2D geographic space into square cells, with each cell size at $25m$. The cell size is selected the same as the evaluation metric for compatibleness, so that the correctly classified images are also the correctly localized images according to our evaluation metric. We remove the Geo-classes which do not contain images, resulting in $1637$ Geo-classes. We append a fully connected layer (random initialization, with weights at $0.01\times randn$) and Softmax-log-loss layer after the NetVLAD pooling layer to predict which class the image belongs to. 

\paragraph{\textbf{SARE loss}}
For our methods (\textit{Our-Ind.}, and \textit{Our-Joint} ), \textit{Our-Ind.} treats multiple negative images independently while \textit{Our-Joint} treats multiple negative images jointly. The two methods only differ in the loss function and gradients computation. For each method, the corresponding gradients are back-propagated to train the CNN.

\paragraph{\textbf{Triplet angular loss}}
For the triplet angular loss \cite{Wang_2017_ICCV}, we use the N-pair loss function (Eq. (8) in their paper) with $\alpha = 45^{\circ}$ as it achieves the best performance on the Stanford car dataset. 

\paragraph{\textbf{N-pair loss}}
For the N-pair loss \cite{sohn2016improved}, we use the N-pair loss function (Eq. (3) in their paper). 

\paragraph{\textbf{Lifted structured loss}}
For the lifted structured loss \cite{oh2016deep}, we use the smooth loss function (Eq. (4) in their paper). Note that training images producing a zero loss ($\tilde{J}_{i,j} < 0$) are pruned out.

\paragraph{\textbf{Ratio loss}}
For the Ratio loss \cite{hoffer2015deep}, we use the MSE loss function since it achieves the best performance in there paper.

\begin{table}[]
\centering
\caption{Datasets  used  in  experiments. The Pitts250k-train dataset is only used to train the Geographic classification CNN \cite{Vo_2017_ICCV}. For all the other CNNs, Pitts30k-train dataset is used to enable fast training. }
\label{tab:datasets}
\begin{tabular}{|l|c|c|}
\hline
Dataset         & \#database images & \#query images \\ \hline
Pitts250k-train & 91,464            & 7,824          \\ \hline
Pitts250k-val   & 78,648            & 7,608          \\ \hline
Pitts250k-test  & 83,952            & 8,280          \\ \hline
Pitts30k-train  & 10,000            & 7,416          \\ \hline
Pitts30k-val    & 10,000            & 7,608          \\ \hline
Pitts30k-test   & 10,000            & 6,816          \\ \hline
TokyoTM-val     & 49,056            & 7,186          \\ \hline
Tokyo 24/7      & 75,984         & 315            \\ \hline   
Sf-0            & 610,773           & 803            \\ \hline
Oxford 5k       & 5063              & 55             \\ \hline
Paris 6k        & 6412              & 220            \\ \hline
Holidays        & 991               & 500            \\ \hline
\end{tabular}
\end{table}

\section{Additional Results}\label{sec::Additional_Results}
\begin{figure*}
\centering
\includegraphics[width=0.99\textwidth,height=\textwidth]{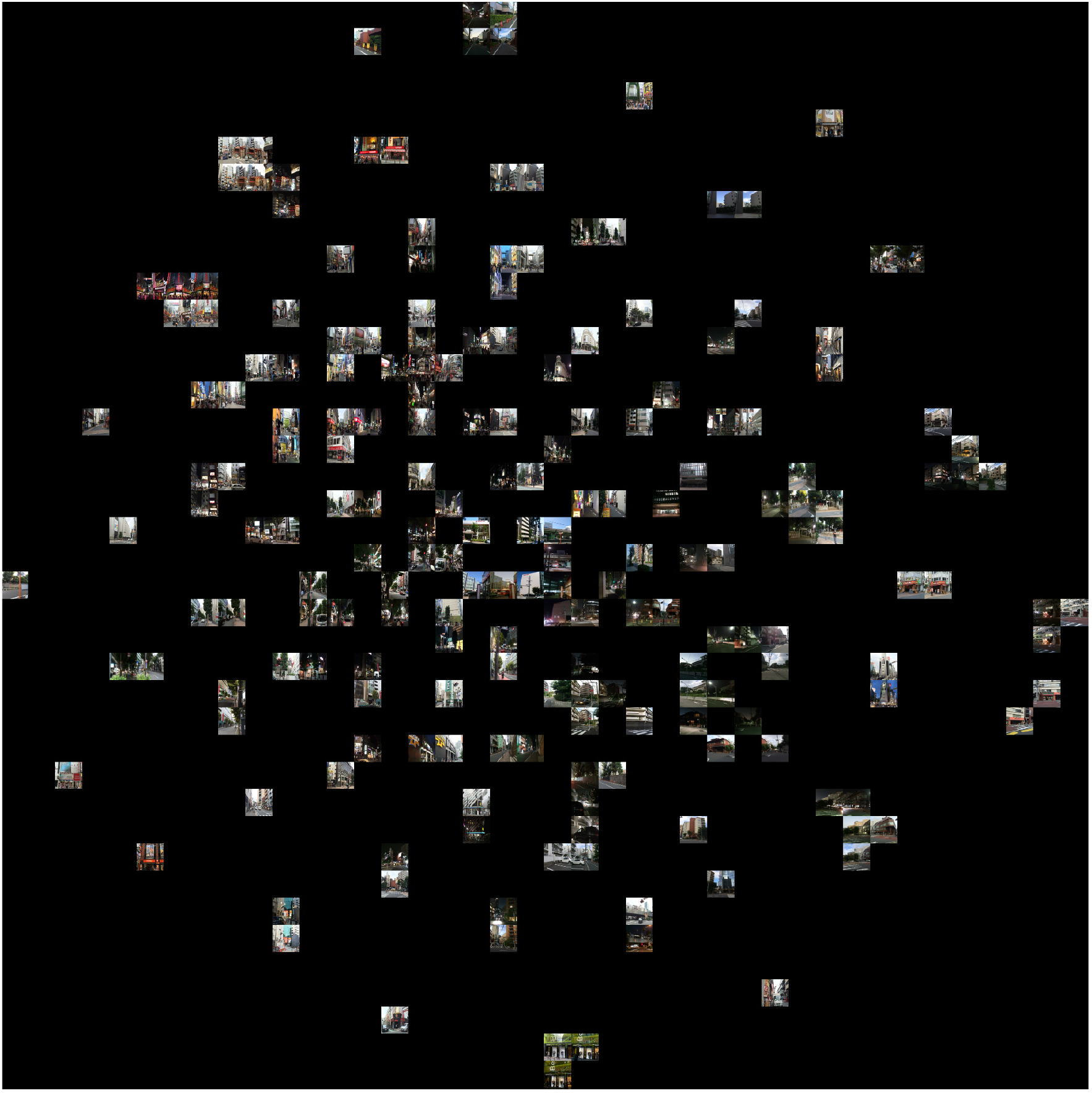}
\caption{{ Visualization of feature embeddings computed by our method ( \textit{Our-Ind.} ) using t-SNE \cite{maaten2008visualizing} on the 24/7 Tokyo-query dataset. (Best viewed in color on screen)}}
\label{fig:embedding_247Tokyo_query}
\end{figure*}

\begin{figure*}
\centering
\includegraphics[width=0.99\textwidth,height=\textwidth]{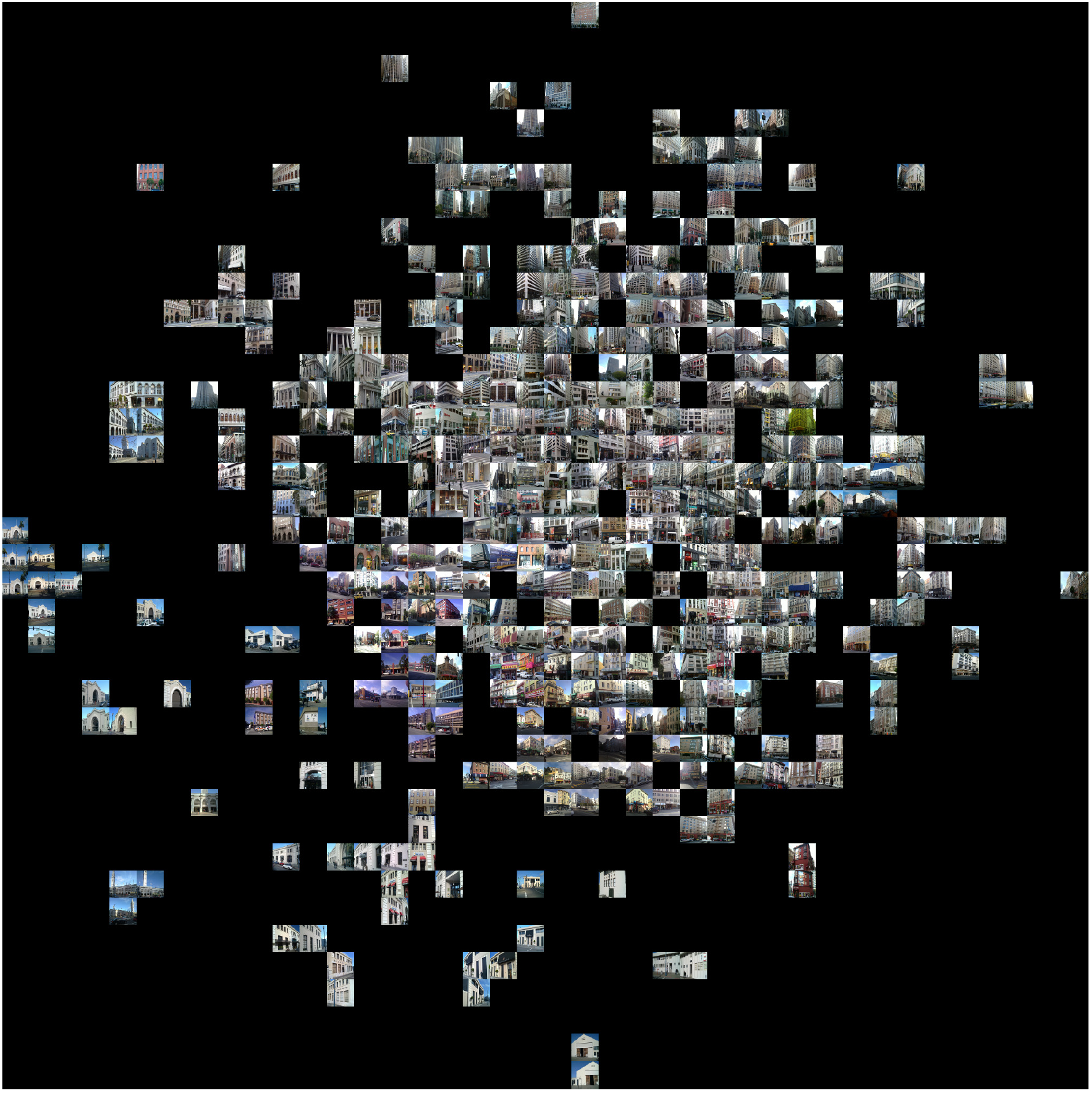}
\caption{{ Visualization of feature embeddings computed by our method ( \textit{Our-Ind.} ) using t-SNE \cite{maaten2008visualizing} on the Sf-0-query dataset. (Best viewed in color on screen)}}
\label{fig:embedding_sf0_query}
\end{figure*}

\paragraph{\textbf{Dataset.}}
Table~\ref{tab:datasets} gives the details of datasets used in our experiments.
\paragraph{\textbf{Visualization of feature embeddings.}}
Fig.~\ref{fig:embedding_247Tokyo_query} and  Fig.~\ref{fig:embedding_sf0_query} visualize the feature embeddings of the  24/7 Tokyo-query and Sf-0-query dataset computed by our method (\textit{Our-Ind.}) in 2-D using the t-SNE \cite{maaten2008visualizing}, respectively. Images are displayed exactly at their embedded locations. Note that images taken from the same place are mostly embedded to nearby 2D positions although they differ in lighting and perspective.

\begin{table*}[]
\centering
\caption{Comparison of Recalls  on  the Pitts250k-test,  TokyoTM-val,  24/7 Tokyo and Sf-0 datasets.}
\label{tab:metric-learning}
\begin{tabular}{|l|c|c|c|c|c|c|c|c|c|c|c|c|}
\hline
\multirow{2}{*}{\backslashbox{Method}{Dataset}} & \multicolumn{3}{c|}{Pitts250k-test} & \multicolumn{3}{c|}{TokyoTM-val} & \multicolumn{3}{c|}{24/7 Tokyo} & \multicolumn{3}{c|}{Sf-0} \\ \cline{2-13} 
                        & r@1        & r@5        & r@10      & r@1       & r@5       & r@10     & r@1       & r@5      & r@10     & r@1     & r@5    & r@10   \\ \hline
Our-Ind.                & \textbf{88.97}      & \textbf{95.50}      & \textbf{96.79}     & 94.49     & 96.73     & 97.30    & 79.68     & 86.67    & 90.48    & \textbf{80.60}   & \textbf{86.70}  & \textbf{89.01} \\ \hline
Our-Joint               & 88.43      & 95.06      & 96.58     & {94.71}     & {96.87}     & \textbf{97.51}    & {80.63}     & {87.30}    & {90.79}    & 77.75   & 85.07  & 87.52  \\ \hline
Contrastive \cite{radenovic2016cnn}              & 86.33      &  94.09      & 95.88     & {93.39}     & {96.09}     & 96.98    & {75.87}     & {86.35}    & {88.89}    & 74.63   & 82.23  & 84.53  \\ \hline
N-pair \cite{sohn2016improved}              & 87.56      & 94.57      & 96.21     & {94.42}     & {96.73}     & 97.41    & {80.00}     & \textbf{89.52}    & \textbf{91.11}    & 76.66  & 83.85  & 87.11  \\ \hline
Angular \cite{Wang_2017_ICCV}              & 88.60      & 94.86      & 96.44     & \textbf{94.84}     & {96.83}     & 97.45    & \textbf{80.95}     & {87.62}    & {90.16}    & 79.51   & 86.57  & 88.06  \\ \hline
Liftstruct  \cite{oh2016deep}             & 87.40      & 94.52      & 96.28     & {94.48}     & \textbf{96.90}     & 97.47    & {77.14}     & {86.03}    & {89.21}    & 78.15   & 84.67  & 87.11  \\ \hline
Geo-Classification  \cite{Vo_2017_ICCV}             & 83.19      & 92.67      & 94.59     & {93.54}     & {96.80}     & 97.50      & {71.43}    & {82.22}    & 85.71   & 67.84  & 78.15 & {81.41}  \\ \hline

Ratio \cite{hoffer2015deep}             & 87.28      & 94.25      & 96.07     & {94.24}     & {96.84}     & 97.41      & {80.32}    & {87.30}    & 88.89   & 76.80  & 85.62 & {87.38}  \\ \hline

\end{tabular}
\end{table*}

\paragraph{\textbf{Metric learning methods}} Table \ref{tab:metric-learning} gives the complete \textit{Recall@N }performance for different methods. Our method outperforms the contrastive loss \cite{radenovic2016cnn} and Geo-classification loss \cite{Vo_2017_ICCV}, while remains comparable with other state-of-the-art metric learning methods. 

\paragraph{\textbf{Image retrieval for varying dimensions.}} Table \ref{tab:retrieval_netvlad_full} gives the comparison of image retrieval performance for different output dimensions.

\begin{table*}[]
\setlength{\tabcolsep}{4pt}
\centering
\caption{Retrieval performance of CNNs trained on Pitts250k-test dataset on image retrieval benchmarks. No spatial re-ranking, or query expansion are performed. The accuracy is measured by the mean Average Precision (mAP).}
\label{tab:retrieval_netvlad_full}
\begin{tabular}{|l|c|c|c|c|c|c|}
\hline
\multirow{2}{*}{Method} & \multirow{2}{*}{Dim.} & \multicolumn{2}{c|}{Oxford 5K}  & \multicolumn{2}{c|}{Paris 6k}   & \multirow{2}{*}{Holidays} \\ \cline{3-6}
                        &                       & full           & crop           & full           & crop           &                           \\ \hline
Our-Ind.                & 4096                  & \textbf{71.66} & \textbf{75.51} & \textbf{82.03} & 81.07          & 80.71                     \\ \hline
Our-Joint               & 4096                  & 70.26          & 73.33          & 81.32          & \textbf{81.39} & \textbf{84.33}            \\ \hline
NetVLAD \cite{arandjelovic2016netvlad}                & 4096                  & 69.09          & 71.62          & 78.53          & 79.67          & 83.00                     \\ \hline
CRN  \cite{kim2017crn}                   & 4096                  & 69.20          & -              & -              & -              & -                         \\ \hline
Our-Ind.                & 2048                  & \textbf{71.11} & \textbf{73.93} & \textbf{80.90} & 79.91          & 79.09                     \\ \hline
Our-Joint               & 2048                  & 69.82          & 72.37          & 80.48          & \textbf{80.49} & \textbf{83.17}            \\ \hline
NetVLAD \cite{arandjelovic2016netvlad}                & 2048                  & 67.70          & 70.84          & 77.01          & 78.29          & 82.80                     \\ \hline
CRN  \cite{kim2017crn}                   & 2048                  & 68.30          & -              & -              & -              & -                         \\ \hline
Our-Ind.                & 1024                  & \textbf{70.31} & \textbf{72.20} & \textbf{79.29} & \textbf{78.54} & 78.76                     \\ \hline
Our-Joint               & 1024                  & 68.46          & 70.72          & 78.49          & 78.47          & \textbf{83.15}            \\ \hline
NetVLAD  \cite{arandjelovic2016netvlad}               & 1024                  & 66.89          & 69.15          & 75.73          & 76.50          & 82.06                     \\ \hline
CRN \cite{kim2017crn}                    & 1024                  & 66.70          & -              & -              & -              & -                         \\ \hline
Our-Ind.                & 512                   & \textbf{68.96} & \textbf{70.59} & \textbf{77.36} & 76.44          & 77.65                     \\ \hline
Our-Joint               & 512                   & 67.17          & 69.19          & 76.80          & \textbf{77.20} & \textbf{81.83}            \\ \hline
NetVLAD  \cite{arandjelovic2016netvlad}               & 512                   & 65.56          & 67.56          & 73.44          & 74.91          & 81.43                     \\ \hline
CRN   \cite{kim2017crn}                  & 512                   & 64.50          & -              & -              & -              & -                         \\ \hline
Our-Ind.                & 256                   & \textbf{65.85} & 67.46          & \textbf{75.61} & 74.82          & 76.27                     \\ \hline
Our-Joint               & 256                   & 65.30          & \textbf{67.51} & 74.50          & \textbf{75.32} & \textbf{80.57}            \\ \hline
NetVLAD  \cite{arandjelovic2016netvlad}               & 256                   & 62.49          & 63.53          & 72.04          & 73.47          & 80.30                     \\ \hline
CRN   \cite{kim2017crn}                  & 256                   & 64.20          & -              & -              & -              & -                         \\ \hline
Our-Ind.                & 128                   & \textbf{63.75} & \textbf{64.71} & \textbf{71.60} & \textbf{71.23} & 73.57                     \\ \hline
Our-Joint               & 128                   & 62.92          & 63.63          & 69.53          & 70.24          & 77.81                     \\ \hline
NetVLAD  \cite{arandjelovic2016netvlad}               & 128                   & 60.43          & 61.40          & 68.74          & 69.49          & \textbf{78.65}            \\ \hline
CRN \cite{kim2017crn}                    & 128                   & 61.50          & -              & -              & -              & -                         \\ \hline
\end{tabular}
\end{table*}

\end{document}